\documentclass[nohyperref]{article}

\usepackage{microtype}
\usepackage{graphicx}
\usepackage{subcaption}
\usepackage{booktabs} 
\usepackage{multirow}
\usepackage{hyperref}

 \usepackage[accepted]{icml2022}

\usepackage{amsmath}
\usepackage{amssymb}
\usepackage{mathtools}
\usepackage{amsthm}
\usepackage{color}
\newcommand{\red}[1]{\textcolor{black}{#1}}

\usepackage[capitalize,noabbrev]{cleveref}

\theoremstyle{plain}
\newtheorem{theorem}{Theorem}

\newtheorem{lemma}{Lemma}

\theoremstyle{definition}
\newtheorem{definition}{Definition}

\theoremstyle{remark}
\newtheorem{remark}{Remark}

\usepackage[textsize=tiny]{todonotes}

\icmltitlerunning{Communication-Efficient Personalized Federated learning via Decentralized Sparse Training}

\begin{document}

\twocolumn[
\icmltitle{DisPFL: Towards Communication-Efficient Personalized Federated Learning \\ via Decentralized Sparse Training}



\icmlsetsymbol{equal}{*}


\begin{icmlauthorlist}
\icmlauthor{Rong Dai}{sch}
\icmlauthor{Li Shen}{comp}
\icmlauthor{Fengxiang He}{comp}
\icmlauthor{Xinmei Tian}{sch,comp2}
\icmlauthor{Dacheng Tao}{comp}
\end{icmlauthorlist}

\icmlaffiliation{sch}{University of Science and Technology of China, Hefei, China}
\icmlaffiliation{comp}{JD Explore Academy, Beijing, China}
\icmlaffiliation{comp2}{Institute of Artificial Intelligence, Hefei Comprehensive National Science Center, Hefei, China}

\icmlcorrespondingauthor{Xinmei Tian}{xinmei@ustc.edu.cn}
\icmlcorrespondingauthor{Li Shen}{mathshenli@gmail.com}

\icmlkeywords{Machine Learning, ICML}

\vskip 0.3in
]



\printAffiliationsAndNotice{}  


\begin{abstract}

Personalized federated learning is proposed to handle the data heterogeneity problem amongst clients by learning dedicated tailored local models for each user. However, existing works are often built in a centralized way, leading to high communication pressure and high vulnerability when a failure or an attack on the central server occurs. In this work, we propose a novel personalized federated learning framework in a decentralized (peer-to-peer) communication protocol named Dis-PFL, which employs personalized sparse masks to customize sparse local models on the edge. To further save the communication and computation cost, we propose a decentralized sparse training technique, which means that each local model in Dis-PFL only maintains a fixed number of active parameters throughout the whole local training and peer-to-peer communication process. Comprehensive experiments demonstrate that Dis-PFL significantly saves the communication bottleneck for the busiest node among all clients and, at the same time, achieves higher model accuracy with less computation cost and communication rounds. Furthermore, we demonstrate that our method can easily adapt to heterogeneous local clients with varying computation complexities and achieves better personalized performances.
\end{abstract}
 
\section{Introduction}


Training deep neural networks is known to be data hungry, but data nowadays are often generated on the edge of the increasingly widely-used mobile devices and Internet of Things (IoT) devices \cite{khan2021federated,nguyen2021federated}. Due to growing concerns about data privacy \cite{voigt2017eu}, sending their local data to a centralized device is usually prohibited. On this ground, federated learning (FL) \cite{mcmahan2017communication}, a distributed training paradigm, becomes a promising privacy-preserving training method that enables a number of clients to produce a global model without sharing local data by aggregating locally trained parameters. One major challenge of FL is the data heterogeneity problem, which means the distributions among clients may vary to a large extent. Personalized FL was thus proposed to achieve personalized individual models for each user through federating instead of using the global model to alleviate this problem.

\begin{figure*}[tb]
 \begin{center}
  \begin{subfigure}{0.22\linewidth}
  \vspace{0.20in}
   \includegraphics[width=1\linewidth]{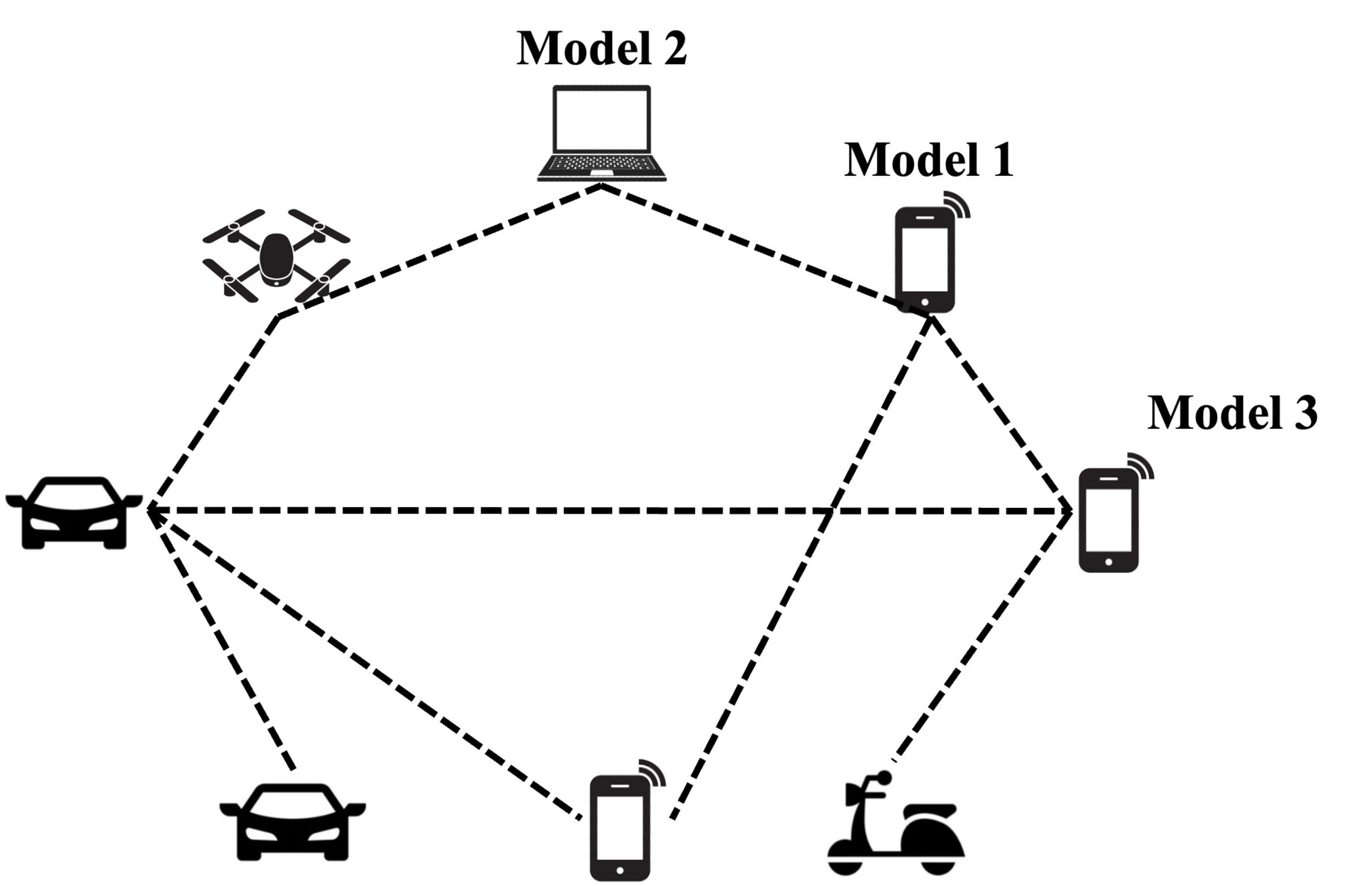}
   \vspace{0.15in}
    \subcaption{}
  \end{subfigure}
  \hspace{1pt}
  \begin{subfigure}{0.37\linewidth}
   \includegraphics[width=1\linewidth]{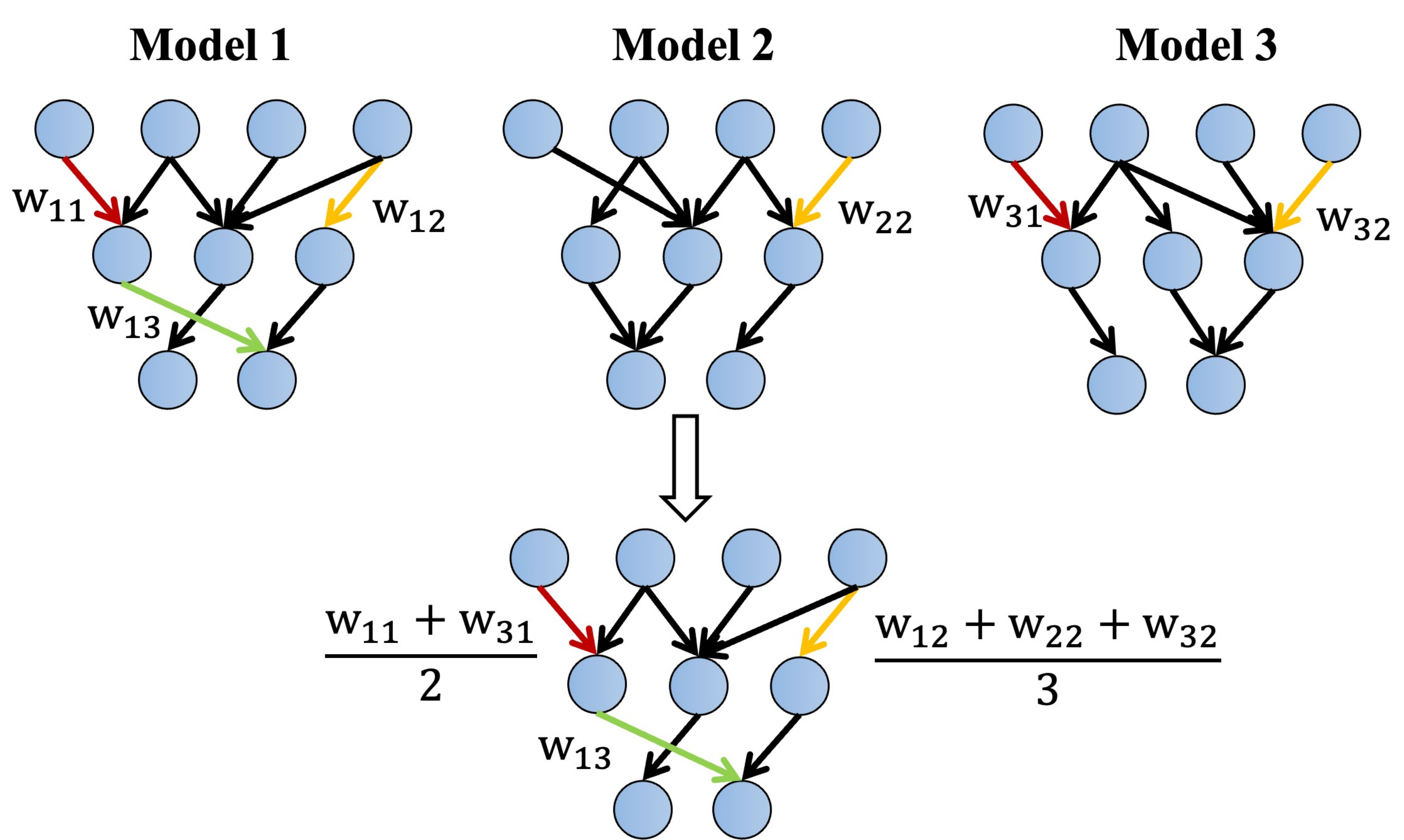}
   \subcaption{}
  \end{subfigure}
  \hspace{1pt}
  \begin{subfigure}{0.37\linewidth}
   \includegraphics[width=1\linewidth]{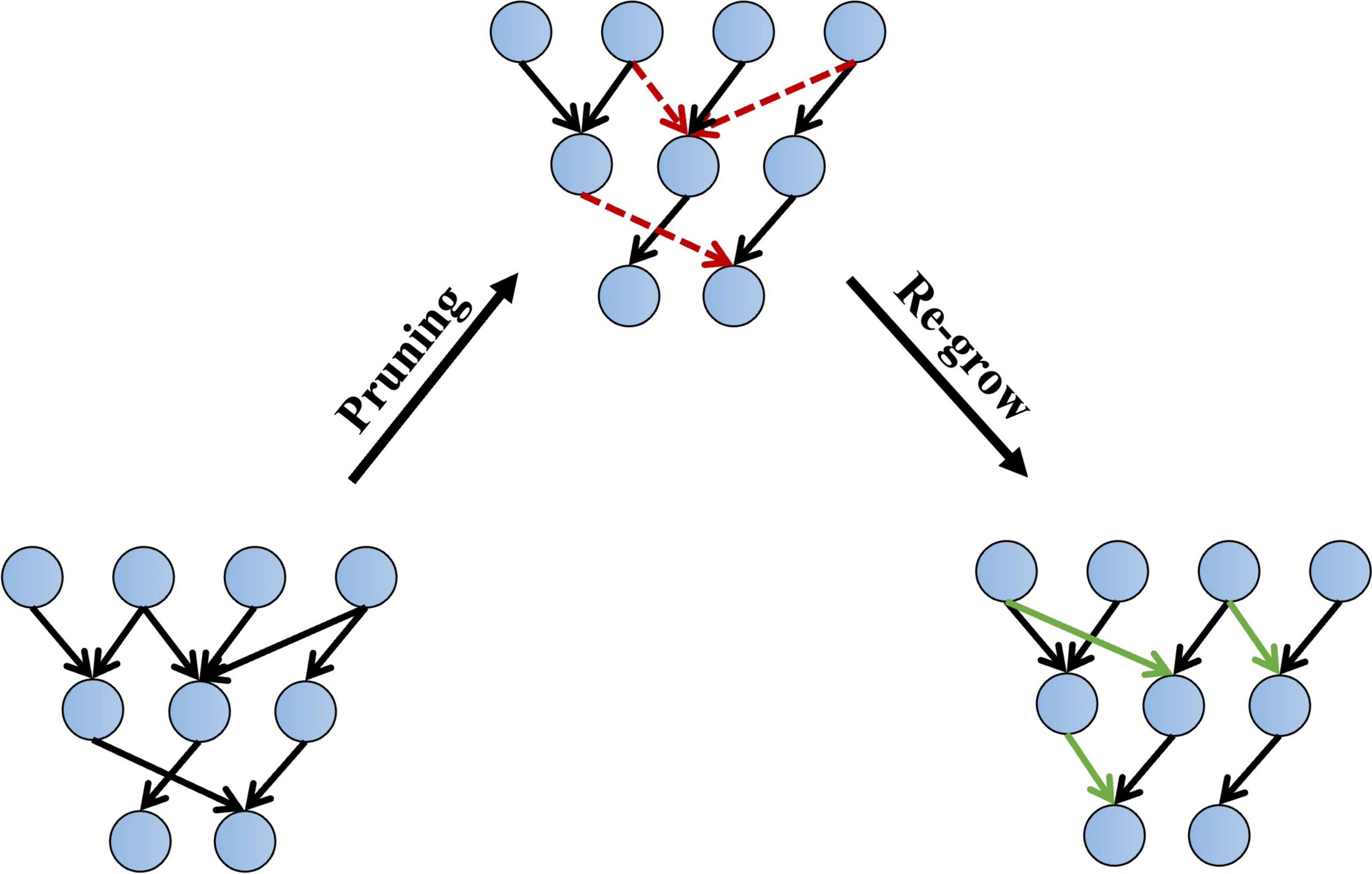}
   \subcaption{}
  \end{subfigure}
 \end{center}
 \vspace{-0.4cm}
 \caption{Overview of Dis-PFL. (a) represents possibly heterogeneous clients in the decentralized federating system, (b) denotes the modified gossip average process by weighted average only the intersection weights, and (c) denotes the local mask searching process.}
 \label{fig:demo}
 \vskip -0.2in
\end{figure*}

Recent works tackling the personalization problems mainly focus on the centralized FL setting \cite{li2021ditto, zhang2020personalized}, where a central server orchestrates the learning amongst clients and is responsible for parameter aggregation after receiving locally trained models on the edge as depicted in Figure \ref{fig:topology}(a). However, this classical FL scheme has a major drawback for the need of the central server. In practice, the central server could face system failure or be maliciously attacked, which may pose a threat to leak users' privacy or jeopardize the training process. Moreover, the communication process all happens on the server-client side, which may cost a quite large communication burden for the server \cite{lian2017can}. With this regard, decentralized FL has recently emerged as a promising method for reducing the communication bandwidth of the busiest node and embracing peer-to-peer communication for faster convergence \cite{lalitha2018fully, sun2021decentralized,licvpr}. In decentralized FL, no global model state exists as in Figure \ref{fig:topology}(b-d), the participating clients follow a communication protocol to reach a so-called consensus model.


 In this paper, we take a further step towards personalization in the decentralized FL setting. Prior works have shown that it's promising to use sparse masks to model personalization for each user \cite{li2020lotteryfl, vahidian2021personalized}, which can also help reduce the communication cost and save computation overhead. However, both methods deal with the centralized setting, and they both employ the technique of dense-to-sparse training. In other words, they must train a dense model on the edge devices at first and then prune it as communication progresses. \red{As a result, these methods can not adapt to client hardware-constrained settings since the model size is restricted to the client capacity.} Moreover, in practice, it is normal to see heterogeneous clients equipped with very different computation, storage and communication capabilities to take part in a same federating system, as shown in Figure \ref{fig:demo} (a). Thus it is crucial to answer the question of how to efficiently use heterogeneous resources to boost the decentralized personalized federating algorithms.

 
To tackle the data heterogeneity and client heterogeneity problem together, we propose Dis-PFL, a {\bf\underline{D}}ecentral{\bf \underline{i}}zed {\bf \underline{s}}parse training based {\bf \underline{P}}ersonnalized {\bf \underline{F}}ederated {\bf \underline{L}}earning approach, to solve the personalized FL problem using customized masks in the decentralized setting by integrating a newly designed decentralized sparse training technique. Instead of deploying the consensus model for each user, Dis-PFL allows each client to own their personalized unique sparse models masked by the tailored mask, allowing them to better adapt to their local data. Specifically, the decentralized sparse training technique mainly consists of three steps: first, weighted average only using the intersection weights of the received neighbor's models as shown in Figure \ref{fig:demo} (b), second, local sparse training with a fixed sparse mask, and third, using gradient information to adjust the current mask for better personalization as depicted in Figure \ref{fig:demo} (c). 
Since all the local models in Dis-PFL are sparse models, the communication cost between peers is greatly reduced. Besides, we show the proposed Dis-PFL can easily adapt to heterogeneous clients equipped with a wide range of computing, memory, and communication capabilities. Empirically, in two classical non-IID settings, we demonstrate that Dis-PFL increases the averaged test accuracy on local test data, reduces the communication cost of the busiest node among all clients, lowers the local computation cost, and requires fewer communication rounds to reach the same target.


To this end, we summarize our contributions as four-fold: 
{\bf (i)} We formulate the personalized federated learning problem using personalized masks and propose Dis-PFL to solve it in a decentralized communication setting.
{\bf (ii)}  The newly proposed decentralized sparse training technique can better aggregate the information, save the local computing, reduce the communication cost and achieve better personalization for decentralized FL.
{\bf (iii)}  Experimental results demonstrate the superiority of the proposed Dis-PFL compared with various baselines and the ability of adaptation to the heterogeneous resources constrained settings.
{\bf (iv)}  A discussion of the sparsity ratio is provided both theoretically and experimentally to analyze the effect of the sparse masks on the generalization ability.


\section{Related Work}
\textbf{Personalized Federated Learning (PFL).}
FL focuses on making the global model more robust to the non-IID distributions by regularizing the local objectives with proximal terms \cite{li2020federated, acar2021federated}, modifying the model aggregation processes \cite{lin2020ensemble, fraboni2021clustered, chen2020fedbe, wang2019federated, zhang2022fine}, client selection \cite{nishio2019client,huang2022stochastic}, etc.
While PFL targets at producing personalized models for each node. There are five primary categories of methods: (1) local fine-tuning \cite{fallah2020personalized, jiang2019improving, cheng2021fine}; (2) adding regularization term \cite{li2021ditto, t2020personalized}; (3) layer personalization \cite{arivazhagan2019federated, collins2021exploiting, liang2020think}; (4) model interpolation \cite{deng2020adaptive, shamsian2021personalized} and (5) model compression \cite{li2020lotteryfl, vahidian2021personalized,huang2022achieving}. Communication cost is one of the major bottlenecks for both FL and PFL, existing works focus on improving the communication efficiency by reducing the volume of transmitted data (i.e., gradients or weights) and can be categorized into three classes: (1) quantization methods \cite{alistarh2017qsgd, wen2017terngrad, yu2019exploring,chen2021quantized}; (2) sparsification \cite{ivkin2019communication, li2021fedmask, bibikar2021federated,chen2020efficient} and (3) hybrid methods \cite{basu2019qsparse, lim20193lc}.

\textbf{Decentralized Learning.} Instead of presuming a central server, decentralized learning targets the same consensus model through peer-to-peer communication. Under assumptions of the topology like doubly stochastic mixing-weights \cite{jiang2017collaborative}, when combing gossip-averaging \cite{blot2016gossip} with SGD, all local models can be proved to converge to a so-called consensus model \cite{lian2017can,chen2021communication}. To tackle the performance degradation problem related to the non-IID scenarios, \citet{lin2021quasi} modify the momentum term to be adaptive to heterogeneous data; \citet{hsieh2020non} replace batch normalization with layer normalization. These methods require to communicate and aggregate the local updates frequently (each iteration), communication rounds thus become a bottleneck. Decentralized federated learning is thus proposed to take benefit of the more local training steps and communicate in a peer-to-peer environment \cite{lalitha2018fully, lalitha2019peer, sun2021decentralized}. Federated schemes have also been extended to time-varying connected communication protocols \cite{warnat2021swarm, yuan2021defed}. \citet{zhu2022topology} prove generalization bounds of decentralized SGD, which suggests the generalization of different topologies is ranked as follows: fully-connected $>$ exponential $>$ grid $>$ ring.

\textbf{Sparse Neural Networks.} Over-parameterization has been shown to be crucial to the dominating performance of deep neural networks \cite{allen2019convergence, zou2019improved}, while some works discover that a sparse model can sufficiently match the performance of the dense model \cite{han2015learning, frankle2018lottery,liu2022unreasonable}. Methods to generate sparse neural networks can be categorized into two main genres: dense-to-sparse methods and sparse-to-sparse methods. Dense-to-sparse methods train from a pre-training dense model and iteratively prune the model \cite{frankle2018lottery, liu2018rethinking, mostafa2019parameter}. While, sparse-to-sparse training has recently been proved to be a viable strategy for improving training efficiency. Starting with a (random) sparse neural network, this paradigm allows the sparse connectivity to grow dynamically during training \cite{mocanu2018scalable,liu2021sparse, liu2022don}. 

\section{Dis-PFL algorithm}

\begin{figure*}[tb]
 \begin{center}
  \begin{subfigure}{0.19\linewidth}
   \includegraphics[width=1\linewidth]{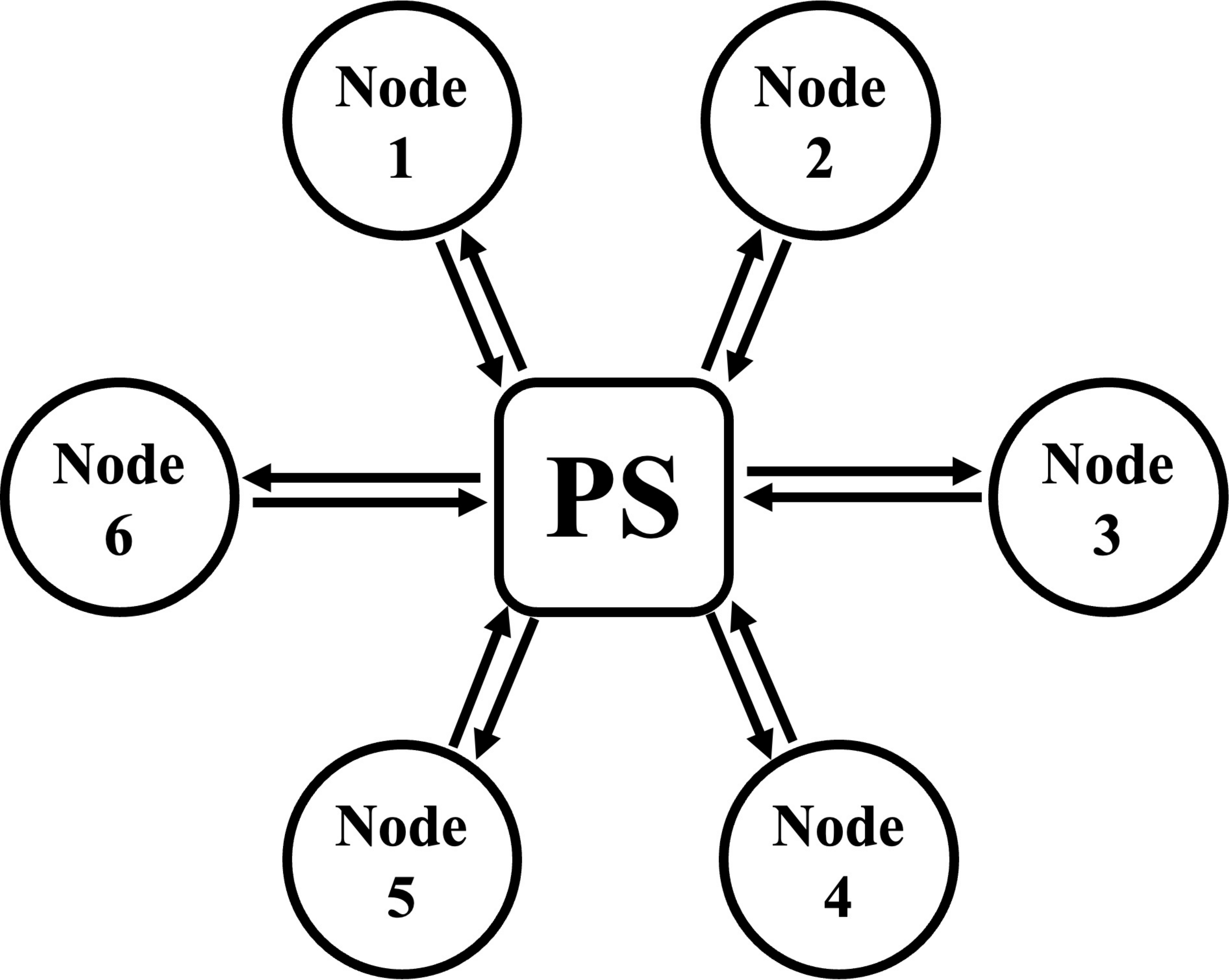}
    \subcaption{}
  \end{subfigure}
  \hspace{1pt}
  \begin{subfigure}{0.19\linewidth}
   \includegraphics[width=1\linewidth]{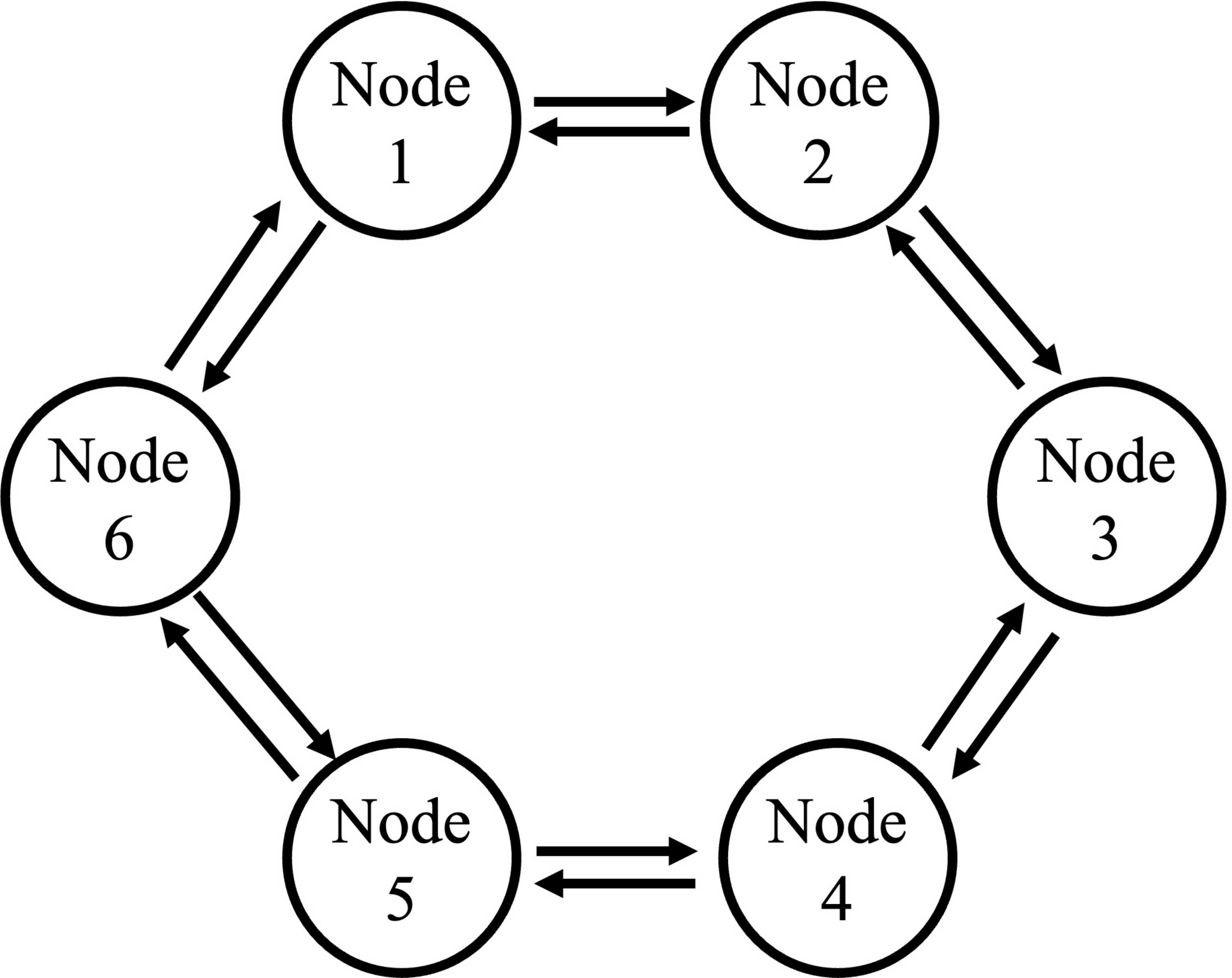}
   \subcaption{}
  \end{subfigure}
  \hspace{1pt}
  \begin{subfigure}{0.19\linewidth}
   \includegraphics[width=1\linewidth]{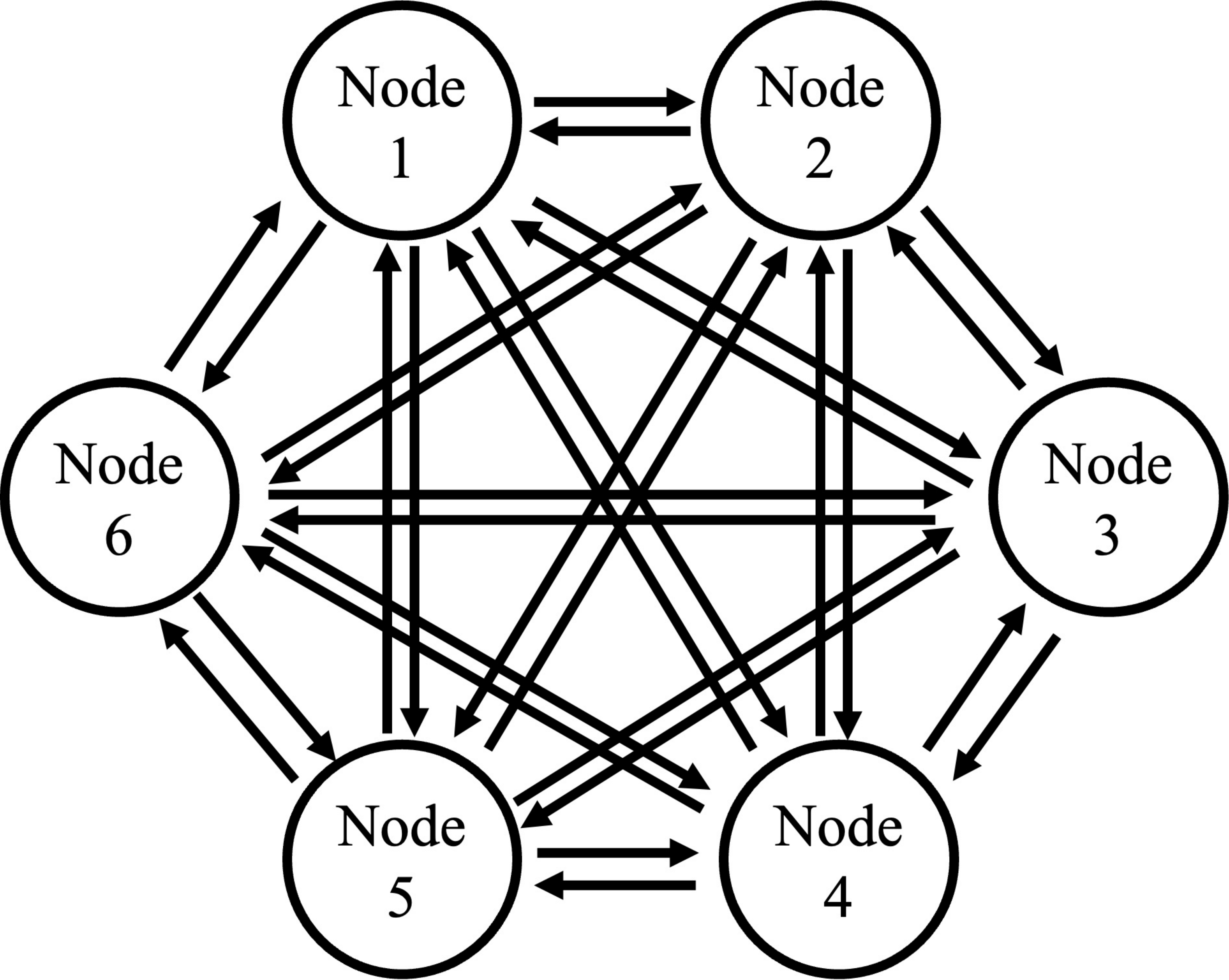}
   \subcaption{}
  \end{subfigure}
   \hspace{1pt}
  \begin{subfigure}{0.38\linewidth}
   \includegraphics[width=1\linewidth]{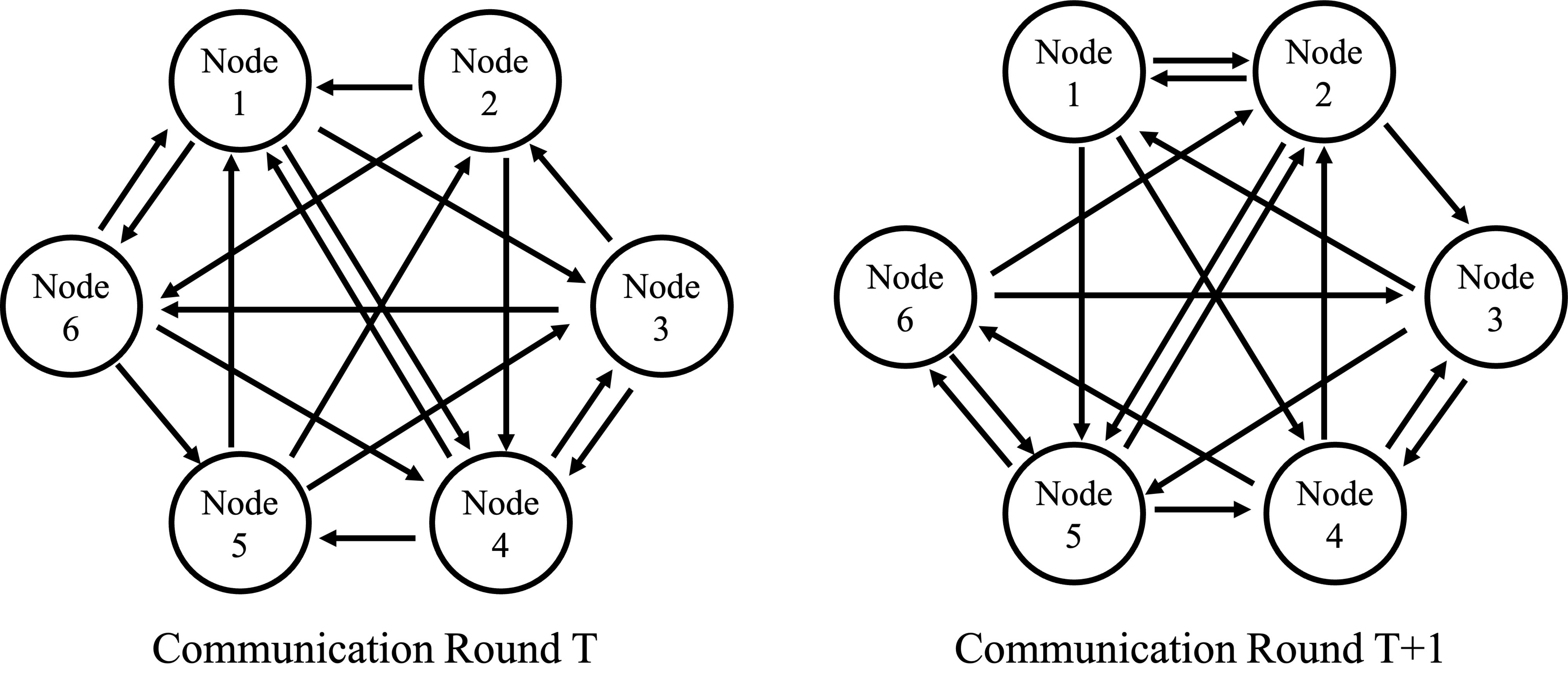}
   \vspace{-0.3in}
   \subcaption{}
  \end{subfigure}
 \end{center}
 \vspace{-0.5cm}
 \caption{Illustrations of various communication protocols, while (a) represents the centralized parameter server network, (b-d) represents the decentralized setting. (b) denotes the ring topology, (c) denotes the fully-connected topology and (d) denotes the time-varying connected topology.}
 \label{fig:topology}
 \vskip -0.1in
\end{figure*}

In this section, we describe the proposed Dis-PFL algorithm. Below, we first present the rigorous formulation of personalized federated learning and derive its variant via utilizing the personalized sparse neural network.  

\subsection{Problem formulation} \label{sec:pf}
We formulate the general PFL problem according to \cite{hanzely2021personalized} into the following optimization task:
\begin{equation}
\begin{gathered}
\small
\min _{\left\{\boldsymbol{w}_{1}, \cdots, \boldsymbol{w}_{K}\right\}} f\left(\boldsymbol{w}_{1}, \cdots, \boldsymbol{w}_{K}\right)=\frac{1}{K} \sum_{k=1}^{K}F_{k}\left(\boldsymbol{w}_{k}\right), \\
F_{k}\left(\boldsymbol{w}_{k}\right):=\mathbb{E}\left[\mathcal{L}_{(\boldsymbol{x}, y) \sim \mathcal{D}_{k}}\left(\boldsymbol{w}_{k} ;(\boldsymbol{x}, y)\right)\right],
\end{gathered}
\label{general-PFL}
\end{equation}
where $\boldsymbol{w}_{k}$ denotes the personalized model for the $k$-th client, $\mathcal{D}_k$ represents the data distribution in the $k$-th client, $F_k(\cdot)$ is the true (population) risk associated with the local distribution and $\mathcal{L}(\cdot;\cdot)$ is the loss function.

This problem could be tackled separately by individual customers with no contact using empirical risk minimization. This algorithm can be called local training and its target can be written as follows:
\begin{equation}
\begin{gathered}
\small
\min _{\left\{\boldsymbol{w}_{1}, \cdots, \boldsymbol{w}_{K}\right\}} f\left(\boldsymbol{w}_{1}, \cdots, \boldsymbol{w}_{K}\right)=\frac{1}{K} \sum_{k=1}^{K}\hat{F}_{k}\left(\boldsymbol{w}_{k}\right), \\
\hat{F}_{k}\left(\boldsymbol{w}_{k}\right):=\sum_{i=1}^{n_{k}}\mathcal{L}\left(\boldsymbol{w}_{k} ;(\boldsymbol{x_i}, y_i)\right),
\end{gathered}
\end{equation}
where $\hat{F}_k(\cdot)$ is the empirical risk associated with $k$-th client's local data and $n_k$ is the total number of the observed data on $k$-th device.

However, the observed local data for each client is often insufficient for training a model with good generalization performance. To achieve better generalization performance, communication and information sharing among clients are encouraged to boost each client's performance. Regularized PFL algorithm can be seen as a general optimization task for these methods, following \cite{chen2021bridging}, the target can be formulated as follows:
\begin{equation}
\small
    \begin{gathered}
        \min _{\left\{\boldsymbol{w}_{1}, \cdots, \boldsymbol{w}_{K}\right\}} f\left(\boldsymbol{w}_{1}, \cdots, \boldsymbol{w}_{K}\right)=\frac{1}{K} \sum_{k=1}^{K}\hat{F}_{k}\left(\boldsymbol{w}_{k}\right)+\mathcal{R}(\cdot), \\
\hat{F}_{k}\left(\boldsymbol{w}_{k}\right):=\sum_{i=1}^{n_{k}}\mathcal{L}\left(\boldsymbol{w}_{k} ;(\boldsymbol{x_i}, y_i)\right),
    \end{gathered}
    \label{R-PFL}
\end{equation}
where the regularizer $R(\cdot)$ indicates the information sharing among clients. It's worth noting that traditional FL which aims at training one global model is also a specific case of Eq. (\ref{R-PFL}) when the regularizer term can be set as $\boldsymbol{w}_1 = \boldsymbol{w}_2 = \cdots = \boldsymbol{w}_K$. By assuming that all users’ data come from the (roughly) similar distribution, it is expected that the global model enjoys a better generalization accuracy on any user distribution over its domain than the user’s own local model.

Instead, we employ personalized mask into the ultimate PFL problem (\ref{general-PFL}) to customize personalized models for each user, the target formulation can be written as:
\begin{equation}
    \begin{gathered}
    \small
        \min _{{\boldsymbol{w}, \boldsymbol{m_1}, \cdots, \boldsymbol{m}_{K}}} f(\boldsymbol{w}, \boldsymbol{m_k})=\frac{1}{K} \sum_{k=1}^{K}F_{k}(\boldsymbol{w}\odot\boldsymbol{m_k}), \\
F_{k}(\boldsymbol{w}\odot \boldsymbol{m}_{k}):=\mathbb{E}\left[\mathcal{L}_{(\boldsymbol{x}, y) \sim \mathcal{D}_{k}}\left(\boldsymbol{w}\odot \boldsymbol{m}_{k} ;(\boldsymbol{x}, y)\right)\right],
    \end{gathered}
    \label{general-SPFL}
\end{equation}
where $\boldsymbol{m_k}\in \{0,1\}^d$ denotes the personalized sparse binary mask for $k$-th client, $\odot$ denotes the Hadamard product for given two vectors. The element of the mask $\boldsymbol{m_k}$ being 1 means that the weight in the global model is active for $k$-th personalized model, otherwise, remains dormant. Our goal is to find a global model $\boldsymbol{w}$ and individual masks $\boldsymbol{m_k}$ for each client, such that the personalized model in client $k$ can be seen as a small part of the global dense model. 

\subsection{Algorithm}


In this section, we propose the Dis-PFL algorithm to solve problem (\ref{general-SPFL}) in the fully decentralized setting. In Dis-PFL algorithm, the decentralized sparse training technique is proposed to integrate to better fit both the data heterogeneity and client heterogeneity problem. In order to meet the computing, memory, and communication constraints, each client only possesses a sparse model during the whole training process. The overall Dis-PFL algorithm is summarized in Algorithm \ref{alg:algo1} and illustrated in Figure \ref{fig:demo}. 

To begin with, the personalized sparsity is determined by each client's computing, memory, and communication restricts $c_k$, thus leading to different masks. The mask is initialized based on Erdos-Renyi Kernel (ERK) \cite{evci2020rigging}, which assigns higher sparsities to layers with more parameters and lower sparsities to layers with fewer parameters. For each communication round, the local client employs the newly designed averaging method as depicted in Figure \ref{fig:demo}(b) to get the initial model $\boldsymbol{w_{k,t+\frac{1}{2}}}$. More specifically, since all the clients share a common dense structure but have different parts of it, we take weighted-average of the received model on the intersection of the remaining part of each model, then multiply it with local mask $\boldsymbol{m}_{k,t}$. 


To further save the overall communication, we integrate the idea from local SGD to the algorithm, a few rounds of local training is done to get the updated sparse model $\boldsymbol{w}_{k,t+1}$. There are two points worth a mention during the local training phase. (\textbf{i}) Since each client only possesses a sparse model, some coordinates of the model weights have been made zero before doing the forward pass, i.e., not all the parameters have to be involved when calculating $L(\widetilde{\boldsymbol{w}}_{k,t,\tau};\xi_{k,t,\tau})$. This implies that the computation overhead in the forward process is potentially saved. (\textbf{ii}) The stochastic gradient, $g_{k,t,\tau}(\widetilde{\boldsymbol{w}}_{k,t,\tau})$ is again masked by $\boldsymbol{m}_{k,t}$ in the backward process, which indicates the gradient for those sparse coordinates has no need to do a backward pass either, leading to the computation cost reduction.

\begin{algorithm}[tb]
\small
   \caption{Dis-PFL}
   \label{alg:algo1}
\begin{algorithmic}[1]
   \STATE {\bfseries Input:} Total number of clients K; Each client's capacity $c_k$; Total communication rounds T
   \STATE {\bfseries Initialization:} Randomly initialize each client's model $\boldsymbol{w}_{i,0}$ and its mask $\boldsymbol{m}_{i,0}$ according to $c_k$
   \STATE {\bfseries Output:} Personalized local models $\boldsymbol{w}_{k,T}$
   \FOR{$t=0$ to $T-1$ do}
   \FOR{node $k$ in parallel}
   \STATE Receive neighbors' models $\boldsymbol{w}_{j,t}$ and corresponding masks  $\boldsymbol{m}_{j,t}$ from neighborhood set $S_{k,t}$
   \STATE $\boldsymbol{w}_{k,t+\frac{1}{2}}=\left(\frac{\boldsymbol{w}_{k,t}+\sum_{j \in S_{k,t}}{ \boldsymbol{w}_{j,t}}} {\boldsymbol{m}_{k,t} + \sum_{j \in S_{k,t}}{ \boldsymbol{m}_{j,t}} }\right) \odot \boldsymbol{m}_{k, t}$
   \STATE $\widetilde{\boldsymbol{w}}_{k,t,0}=\boldsymbol{w}_{k,t+\frac{1}{2}}$
   \FOR{$\tau=0$ to $N-1$ do}
   \STATE Sample a batch of data $\xi_{k,t,\tau}$ from local dataset
   \STATE $g_{k,t,\tau}(\widetilde{\boldsymbol{w}}_{k,t,\tau})=\bigtriangledown_{\widetilde{\boldsymbol{w}}_{k,t,\tau}} L(\widetilde{\boldsymbol{w}}_{k,t,\tau};\xi_{k,t,\tau})$
   \STATE $\widetilde{\boldsymbol{w}}_{k,t,\tau+1}=\widetilde{\boldsymbol{w}}_{k,t,\tau}-\eta \boldsymbol{m}_{k,t} \odot g_{k,t,\tau}(\widetilde{\boldsymbol{w}}_{k,t,\tau})$
   \ENDFOR
   \STATE $\boldsymbol{w}_{k,t+1}=\widetilde{\boldsymbol{w}}_{k,t,N}$
   \STATE Call Algorithm  \ref{alg:algo2} to get new mask $\boldsymbol{m}_{k,t+1}$
   \ENDFOR
   \ENDFOR
\end{algorithmic}
\end{algorithm}

After doing few steps of local training, Algorithm \ref{alg:algo2} detailed in the appendix is called to get the new mask $\boldsymbol{m_{k,t+1}}$ as depicted in Figure \ref{fig:demo} (c). Inspired from \cite{evci2020rigging}, we take similar steps to update the local mask on each client. The pruning rate $\alpha_t$ is calculated through cosine annealing technique mentioned in \cite{liu2021we}. For each layer, each local client first prunes out $\alpha_t$-proportion of weights with the smallest magnitude and then utilizes the gradient information to recover weights with the highest gradient information, which are expected to perform better personalization for the current client.

Overall, to accommodate to the decentralized PFL context, the decentralized sparse training technique we proposed mainly includes three parts, the modified gossip average process regarding how to get the initial model for each client after receiving neighbor's information, local training with fixed sparse masks process and the local mask searching process regarding how to evolve the personalized mask to better adapt to local client's data. We highlight the contributions we make to extend the traditional sparse training techniques like Rigl \cite{evci2020rigging}, and Set \cite{mocanu2018scalable} to the proposed decentralized sparse training technique as follows: (\textbf{i}) To deal with the data heterogeneity and to learn different sparse models for each client, decentralized sparse training operates on the local client instead of operating on the centralized device. (\textbf{ii}) The information integrating process is delicately designed to best fuse the sparse models with different masks. And it can easily adapt to the client heterogeneity settings. (\textbf{iii}) Traditional dynamic sparse training approaches apply the next mask to the model weight immediately after it is created and assign zero weight to the previous undiscovered parameters. However, as demonstrated by \cite{liu2021we}, recovering a coordinate from 0 to a suitable value may need additional training steps. This issue is automatically alleviated in the decentralized sparse training process, since the value of the recovered coordinates may be obtained using the modified gossip average step, allowing for an appropriate warm-up.

\subsection{Generalization analysis}
Generalization characterizes the performance on unseen data of a well-trained model \cite{mohri2018foundations, he2020recent}. 
Suppose $\tilde D$ is the union distribution of $D_k$. Following notations in Section \ref{sec:pf}, the expected risk and empirical risk of the global model $\boldsymbol{w}$ are defined as:
\begin{gather}
\small
    \mathcal{R} = \mathbb{E}_{\boldsymbol{x}\sim \mathcal{\tilde D}}\mathcal{L}(\boldsymbol{w; \boldsymbol{x}}),~
    \hat{\mathcal{R}}
    =  \frac{1}{K}\sum_{k=1}^K \frac{1}{n_k}\sum_{i=1}^{n_k}\mathcal{L}(\boldsymbol{w; \boldsymbol{x_{i}}}).
\end{gather}

Suppose $\beta_k$ denotes the proportion of the remaining model parameters (the opposite of sparsity ratio) on each client. $\beta$ denotes the proportion of the remaining parameters over the aggregations of all clients' sparse masks, (aka the proportion for model $\boldsymbol{w}$). Then for the hypothesis $\mathcal{A}(S)$,  $\boldsymbol{w}$ learned by Algorithm \ref{alg:algo1} on the training sample set $S$, we obtain a generalization bound as follows:
\begin{theorem}
\label{thm:main}
Suppose the loss function $\Vert \mathcal{L}\Vert_{\infty}\le 1$ and the training sample size $N\ge\frac{2}{\varepsilon'^{2}} \ln \left(\frac{16}{e^{-\varepsilon'}\delta'}\right)$. Then, for any data distribution $\mathcal{\tilde D}$ over data space $\mathcal{Z}$, we have the following inequality,
\begin{equation*}
\small
\mathbb{P}\left[\left|\hat{\mathcal{R}}_S(\mathcal{A}(S)) - \mathcal{R}(\mathcal{A}(S))\right| < 9\varepsilon'\right] > 1-\frac{e^{-\varepsilon'}\delta'}{\varepsilon'} \ln \left(\frac{2}{\varepsilon'}\right).
\end{equation*}
Where \begin{normalsize}
    $\varepsilon'=\sqrt{2T  \log \left( \frac{1}{\tilde\delta}\right)\tilde\varepsilon^2} +T   \tilde\varepsilon \frac{e^{ \tilde\varepsilon}-1}{e^{ \tilde\varepsilon}+1},$
\end{normalsize}
\begin{small}
\begin{align*}
    \delta' = & e^{-\frac{\varepsilon'+T \tilde{\varepsilon}}{2}}\left(\frac{1}{1+e^ {\tilde\varepsilon}}\left(\frac{2T {\tilde{\varepsilon}}}{T {\tilde{\varepsilon}}-\varepsilon'}\right)\right)^T\left(\frac{T {\tilde{\varepsilon}}+\varepsilon'}{T {\tilde{\varepsilon}}-\varepsilon'}\right)^{-\frac{\varepsilon'+T {\tilde{\varepsilon}}}{2 {\tilde{\varepsilon}}}}
\\
&+2-\left(1-e^{ {\tilde{\varepsilon}}}\frac{\delta}{1+e^{ {\tilde{\varepsilon}}}}\right)^{\left \lceil  \frac{\varepsilon'}{ {\tilde{\varepsilon}}}\right \rceil}\left(1-\frac{\delta}{1+e^{ {\tilde{\varepsilon}}}}\right)^{T-\left \lceil  \frac{\varepsilon'}{ {\tilde{\varepsilon}}}\right \rceil} 
 \\
 &- \left(1-\frac{\delta}{1+e^{ {\tilde{\varepsilon}}}}\right)^{T}.
\end{align*}
\end{small}
	\begin{center}
	 \begin{normalsize}
	$\tilde\varepsilon =\log \left(\frac{N-\tau}{N}+\frac{\tau}{N}\exp\left(\frac{\sqrt{2}\beta D_g\sigma\frac{1}{\tau}\sqrt{\log\frac{1}{\delta}}+\frac{1}{\tau^2}\beta^2D_g^2}{2\sigma^2}\right)\right)$,
	 \end{normalsize}
	\end{center}
	 $T$ is the training iterations, $\tau$ is the batch-size, $\sigma$ is the Gaussian noise variance and $D_g$ denotes the maximum of local gradient's diameter, $\delta$ is defined in Eq.(\ref{delta}).
\end{theorem}
The proof is inspired by \cite{he2021tighter}; details are given in Appendix \ref{ap:proof}.

\begin{remark}
The personalized models for each client solved by problem (\ref{general-SPFL}) can be seen as different parts of the global model with different sparsity. Theorem \ref{thm:main} suggests that a more sparse network (with a smaller $\beta_k$, leading to a smaller $\beta$), the term $\tilde\varepsilon$ is smaller, and thus the generalization bound (gap between the training error and test error) is smaller. Therefore, the generalization performance is better.
\end{remark}

\section{Experiments}
In this section, we conduct intensive experiments to verify the efficacy of the proposed Dis-PFL. We leave the detailed implementation details and extended experimental results to the supplementary due to space limitation. Code is available at https://github.com/rong-dai/DisPFL.

\subsection{Experiment Setup}
\paragraph{Dataset and Data partition.} We evaluate the performance of the proposed algorithm on three image classification datasets: \textbf{CIFAR-10}, \textbf{CIFAR-100} \cite{krizhevsky2009learning} and \textbf{Tiny-Imagenet}. We consider two different scenarios for simulating non-identical data distributions across federating clients. \textbf{Dir Partition} following works \cite{hsu2019measuring}, where we partition the training data according to a Dirichlet distribution Dir($\alpha$) for each client and generate the corresponding test data for each client following the same distribution. We specify $\alpha=0.3$ for CIFAR-10, $\alpha=0.2$ for CIFAR-100 and Tiny-Imagenet. We also evaluate with the \textbf{pathological partition} setup similar as in \cite{zhang2020personalized}, in which each client is only assigned limited classes at random from the total number of classes. We specify each client possesses 2 classes for CIFAR-10, 10 classes for CIFAR-100, and 20 classes for Tiny-Imagenet.


\paragraph{Baselines.} We compare our methods with diverse set of baselines both from centralized federated learning and decentralized federated learning. A simple baseline called \textbf{Local} is implemented with each client separately only do local training on their own data. Centralized baselines include \textbf{FedAvg} \cite{mcmahan2017communication}, \textbf{FedAvg-FT}\cite{cheng2021fine}, \textbf{Ditto} \cite{li2021ditto}, \textbf{FOMO} \cite{zhang2020personalized}, and \textbf{SubFedAvg} \cite{vahidian2021personalized}. For decentralized federated learning setting, we take the commonly used \textbf{D-PSGD} \cite{lian2017can} as one baseline. Noticeably, to accommodate to the FL setting, we extend the local training from only one iteration of stochastic gradient descent to several epochs over local data. Further, similarly to \textbf{FedAvg-FT}, we extend it to \textbf{D-PSGD-FT} as another baseline where the final models are acquired by performing few fine-tuning steps on the global consensus model with local data.


\paragraph{Communication protocol.} For centralized FL baselines, the communication happens between the global server and clients. Typically, in each communication round, the server may only receive part of all the clients' information due to the communication bandwidth. Thus, for a fair comparison with centralized FL setting, we apply a dynamic time-changing connection topology for decentralized methods as depicted in Figure \ref{fig:topology}(d), where each client can only communicate with restricted random selected neighbors and may differ among each communication round. We make sure that the connections of the busiest node are no more than the connections of the server, which can be seen as the busiest node in the centralized setting. We also experiment on topology designed for decentralized setting including ring (Figure \ref{fig:topology}(b)) and fully-connected (Figure \ref{fig:topology}(c)).

\begin{table}[t]
\scriptsize
\caption{Table illustrating performance of different methods. \textit{Comm }\red{means the maximum comm cost (download/upload) for the busiest devices(server in centralized settings)} and \textit{FLOPS} denotes \red{the total floating operations for each client during the whole local phase for each communication round.}}
\label{centralized-main-table}
\vskip 0.15in
\centering
\begin{tabular}{cccccc}
\toprule
Task                           & Methods   & \begin{tabular}[c]{@{}c@{}}Dir Part\\ Acc\end{tabular} & \begin{tabular}[c]{@{}c@{}}Path Part\\ Acc\end{tabular} & \begin{tabular}[c]{@{}c@{}}Comm\\ (MB)\end{tabular} & \begin{tabular}[c]{@{}c@{}}FLOP\\ (1e12)\end{tabular} \\ \midrule
\multirow{9}{*} {\rotatebox{90}{CIFAR-10}}     & Local     & 61.55$\pm$0.2                                          & 86.48$\pm$0.2                                           & -                                                    & 8.3                                                      \\
                               & FedAvg    & 78.07$\pm$0.5                                          & 54.53$\pm$0.6                                           & 446.9                                                    & 8.3                                                      \\
                               & FedAvg-FT & 81.20$\pm$0.5                                          & 84.96$\pm$0.2                                           & 446.9                                                    & 8.3                                                      \\
                               & D-PSGD    & 79.02$\pm$0.4                                          & 58.07$\pm$0.5                                           & 446.9                                                    & 8.3                                                       \\
                               & D-PSGD-FT & 83.90$\pm$0.2                                          & 90.87$\pm$0.2                                           & 446.9                                                    & 8.3                                                       \\
                               & Ditto     & 74.68$\pm$0.2                                          & 87.73$\pm$0.1                                           & 446.9                                                    & 8.3                                                      \\
                               & FOMO      & 64.68$\pm$0.2                                          & 88.24$\pm$0.1                                           & 446.9                                                    & 8.3                                                      \\
                               & SubFedAvg & 76.70$\pm$0.2                                          & 88.30$\pm$0.2                                           & 278.8                                                    & \textbf{4.7}                                                      \\
                               & Dis-PFL   & \textbf{85.70$\pm$0.2}                                 & \textbf{91.05$\pm$0.2}                                  & \textbf{223.4}                                                    & 7.0                                                      \\ \midrule
\multirow{9}{*}{\rotatebox{90}{CIFAR-100}}     & Local     & 29.23$\pm$0.2                                          & 52.46$\pm$0.2                                           & -                                                    & 8.3                                                      \\
                               & FedAvg    & 41.72$\pm$0.5                                          & 33.24$\pm$0.6                                           & 448.7                                                    & 8.3                                                      \\
                               & FedAvg-FT & 49.19$\pm$0.5                                          & 63.53$\pm$0.7                                           & 448.7                                                    & 8.3                                                      \\
                               & D-PSGD    & 41.87$\pm$0.4                                          & 35.42$\pm$0.2                                           & 448.7                                                     & 8.3                                                      \\
                               & D-PSGD-FT & 51.42$\pm$0.4                                          & 67.24$\pm$0.1                                           & 448.7                                                    & 8.3                                                      \\
                               & Ditto     & 38.26$\pm$0.2                                          & 54.02$\pm$0.3                                           & 448.7                                                    & 8.3                                                      \\
                               & FOMO      & 28.39$\pm$0.1                                          & 52.74$\pm$0.1                                           & 448.7                                                    & 8.3                                                      \\
                               & SubFedAvg & 43.91$\pm$0.2                                          & 60.67$\pm$0.1                                           & 346.6                                                    & \textbf{5.7}                                                      \\
                               & Dis-PFL   & \textbf{53.48$\pm$0.3}                                 & \textbf{68.64$\pm$0.4}                                  & \textbf{224.3}                                                    & 7.0                                                      \\ \midrule
\multirow{9}{*}{\rotatebox{90}{Tiny-Imagenet}} & Local     & 6.76$\pm$0.2                                                       & 17.68$\pm$0.3                                           & -                                                    & 66.6                                                      \\
                               & FedAvg    & 12.30$\pm$0.3               & 10.40$\pm$0.3                                            & 450.7                                                    & 66.6                                                      \\
                               & FedAvg-FT & 14.80$\pm$0.2        & 28.30$\pm$0.2   & 450.7                                                    & 66.6                                                      \\
                               & D-PSGD    & 12.13$\pm$0.5                                                        &  16.50$\pm$0.4                                            & 450.7                                                    &  66.6                                                     \\
                               & D-PSGD-FT & 15.50$\pm$0.3                                                       & 28.60$\pm$0.3                                           & 450.7                                                    & 66.6                                                      \\
                               & Ditto     & 15.69$\pm$0.2                                                       & 24.55$\pm$0.3                                        & 450.7                                                    & 66.6                                                      \\
                               & FOMO      & 5.20$\pm$0.4                                                       & 9.39$\pm$0.3                                                        & 450.7                                                    &  66.6                                                     \\
                               & SubFedAvg & 12.18$\pm$0.4                                            & 19.73$\pm$0.5                                           & 290.9                                                    & \textbf{40.2}                                                      \\
                               & Dis-PFL   & \textbf{16.95$\pm$0.4}                                                      & \textbf{31.71$\pm$0.4}                                & \textbf{225.3}                                                    & 54.5                                                      \\
                               \bottomrule
\end{tabular}

\vskip -0.1in
\end{table}

\paragraph{Configurations.} Unless otherwise mentioned, all the reported results are averaged for at least two random seeds. The final accuracy is calculated through the average of each local client's test accuracy on their own test data. We choose modified Resnet-18 as the backbone for all three datasets. The total client number is set to 100, and we restrict the busiest node to communicate with at most 10 neighbors. The sparsity of the local model is set 0.5 for all the clients in the main experiments and may vary when it comes to heterogeneous clients' capabilities which will be further explained in the corresponding experiments subsection.

\begin{table}[t]
\caption{Table illustrating performance in different topology (FC means fully-connected) on CIFAR-10. Results on CIFAR-100 and Tiny-Imagenet can be found in the appendix \ref{ap:top}}
\label{decentralized-main-table}
\vskip 0.15in
\centering
\scriptsize
\begin{tabular}{cccccc}
\toprule
\multirow{2}{*}{}        & \multirow{2}{*}{Method} & \multirow{2}{*}{\begin{tabular}[c]{@{}c@{}}Dir Part\\ Acc\end{tabular}} & \multirow{2}{*}{\begin{tabular}[c]{@{}c@{}}Path Part\\ Acc\end{tabular}} & \multirow{2}{*}{\begin{tabular}[c]{@{}c@{}}Comm\\(MB)\end{tabular}} & \multirow{2}{*}{\begin{tabular}[c]{@{}c@{}}FLOPS\\(1e12)\end{tabular}} \\
                                                           &                         &                                                                    &                                                                     &                                                                                   &                                                                                      \\
                              \midrule
 Separate                         & Local                   & 61.66$\pm$0.2                                                                      & 86.48$\pm$0.2                                                               & -                                                                                  & 8.3                                                                                     \\
\midrule
                                \multirow{3}{*}{Ring}            & D-PSGD                  & 49.46$\pm$0.2                                                                   & 24.42$\pm$0.5                                                               & 89.4                                                                                 & 8.3                                                                                     \\
                                                                &  D-PSGD-FT               & 67.80$\pm$0.3                                                                    & 86.68$\pm$0.2                                                      & 89.4                                                                                  & 8.3                                                                                     \\
                                                                & Dis-PFL           & \textbf{67.81$\pm$0.2}                                                                   & \textbf{86.70$\pm$0.2}                                                               & \textbf{44.6}                                                                                  & \textbf{7.0}                                                                                     \\
                               \midrule
                                \multirow{3}{*}{FC} & D-PSGD                  & 79.56$\pm0.2$                                                                   & 60.45$\pm$0.3                                                               & 4423.9                                                                                  &    8.3                                                                                  \\
                                                                & D-PSGD-FT               & 84.57$\pm$0.2                                                                   & 90.58$\pm$0.3                                                               &  4423.9                                                                                  & 8.3                                                                                     \\
                                                                 & Dis-PFL          & \textbf{86.71$\pm$0.2}                                                                   & \textbf{91.14$\pm$0.1}                                                      & \textbf{2211.4}                                                                                  & \textbf{7.0}                                                                                     \\ 

                               \bottomrule
\end{tabular}

\end{table}

\subsection{Main experiments evaluation}

\paragraph{Test accuracy of the personalized model. }
In Table \ref{centralized-main-table} and Table \ref{decentralized-main-table}, we show that our method Dis-PFL achieves remarkable performances and outperforms all the centralized or decentralized baselines by a large margin over all the three datasets and two classic types of non-IID partitions. For the baselines, we notice that those methods targeting a global consensus model without encouraging personalization of local models, e.g. FedAvg and D-PSGD may perform even worse than the simple separate local training baseline. This is due to the fact that the non-IID phenomenon is sometimes harsh for a simple global model to cover all the data. While FedAvg-FT, D-PSGD-FT, Ditto, and FOMO, targeting at learning personalized dense models for each client, perform better than the other baselines but still worse than our methods. Also, we notice some methods work well on pathological non-IID distribution, but may fail to work on Dirichlet distribution. We assume this is attributed to the scarcity of local data for each class, and thus hard for their methods to work. In contrast, our method performs better among others no matter what the partitions are. Our method also outperforms SubFedAvg to a large extent, which uses the same concept of learning a unique mask for each client. 

\begin{figure*}[tb]
\small
 \begin{center}
   \includegraphics[width=0.9\linewidth]{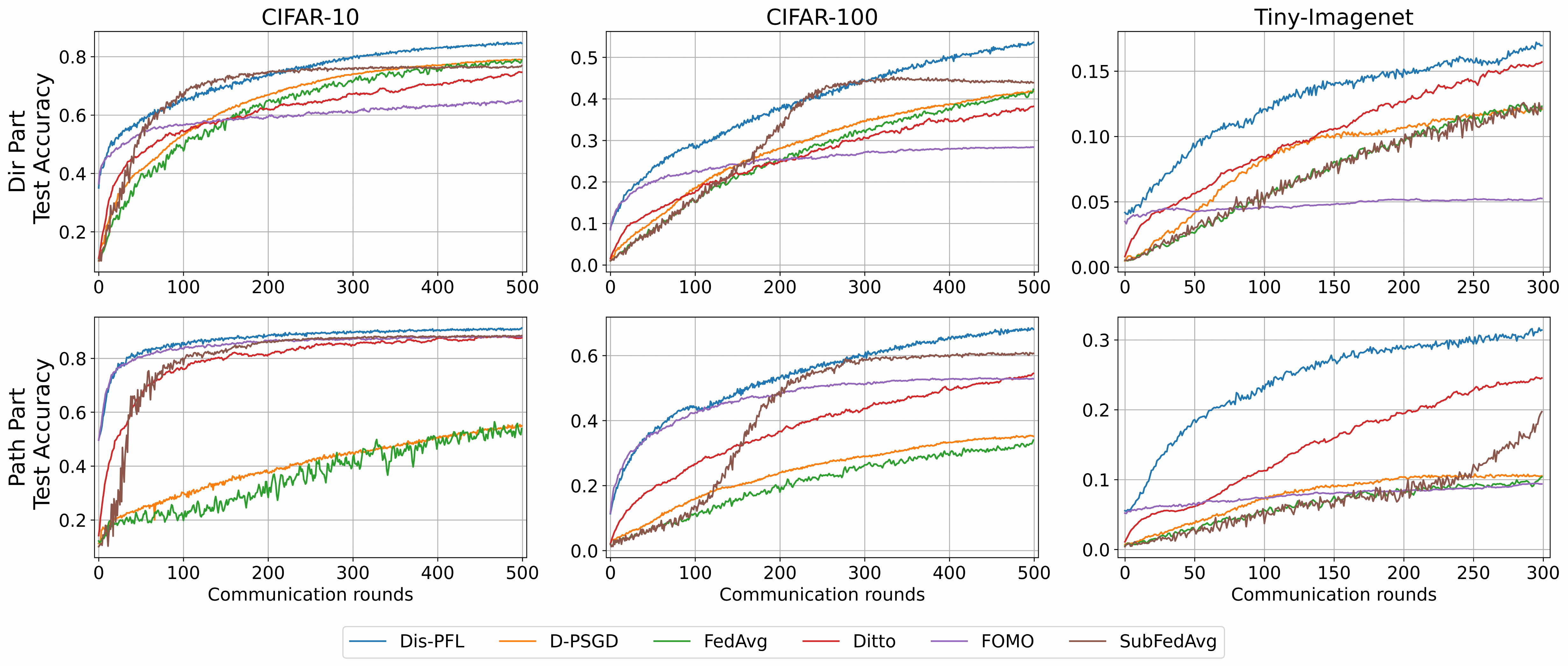}
 \end{center}
 \vspace{-5pt}
 \caption{Learning curves on three Datasets with two data partitions.}
 \label{fig:main_acc}
\end{figure*}

\begin{table*}[tb]
\centering
\scriptsize
\caption{Performance of each method when it comes to the heterogeneous clients with different constraints settings. \textit{Comm} here denotes the averaged communication cost per node for each communication round.}
\label{extended-table}
\vskip 0.15in
\begin{tabular}{ccccc|cccc}
\toprule
                                                                                       & \multicolumn{4}{c|}{Resnet-18}                                                                                                                                                                                                  & \multicolumn{4}{c}{VGG-11}                                                                                                                                                                                                      \\ \cmidrule{2-9}
Methods                                                                                & \begin{tabular}[c]{@{}c@{}}Dir Part\\ Acc\end{tabular} & \begin{tabular}[c]{@{}c@{}}Path Part\\ Acc\end{tabular} & \begin{tabular}[c]{@{}c@{}}Comm\\ (MB)\end{tabular} & \begin{tabular}[c]{@{}c@{}}FLOPS\\ (1e11)\end{tabular} & \begin{tabular}[c]{@{}c@{}}Dir Part\\ Acc\end{tabular} & \begin{tabular}[c]{@{}c@{}}Path Part\\ Acc\end{tabular} & \begin{tabular}[c]{@{}c@{}}Comm\\ (MB)\end{tabular} & \begin{tabular}[c]{@{}c@{}}FLOPS\\ (1e11)\end{tabular} \\ \midrule
D-PSGD (20\% parmas)         & 72.24                                                  & 59.83                                                   & 89.4                                                & 20.9                                                   & 66.00                                                  & 59.60                                                   & 73.8                                                & 5.7                                                    \\
D-PSGD-FT (20\% params)                      & 78.46                                                  & 87.74                                                   & 89.4                                                & 20.9                                                   & 74.71                                                  & 87.76                                                   & 73.8                                                & 5.7                                                    \\
D-PSGD (50\% params)                         & 79.02                                                  & 62.12                                                   & 223.4                                               & 45.8                                                   & 76.71                                                  & 73.60                                                   & 184.6                                               & 12.6                                                   \\
D-PSGD-FT (50\% parmas)          & 83.77                                                  & 88.76                                                   & 223.4                                               & 45.8                                                   & 82.60                                                  & 91.94                                                   & 184.6                                               & 12.6                                                   \\ \midrule
 Dis-PFL (Setting ii)    &  84.33 & 90.84 & 268.0 & 71.3 
 & 83.04 & 91.02 & 221.4  & 18.0                                                  \\
  Dis-PFL (Setting i)
  & \textbf{84.85} & \textbf{91.27}  & 223.4 & 70.4
  & \textbf{84.95}  & \textbf{92.11}  & 184.6 & 17.3 \\                               
                \bottomrule
\end{tabular}
\end{table*}

\begin{figure}[tb]
 \begin{center}
  \begin{subfigure}{0.48\linewidth}
   \includegraphics[width=1\linewidth]{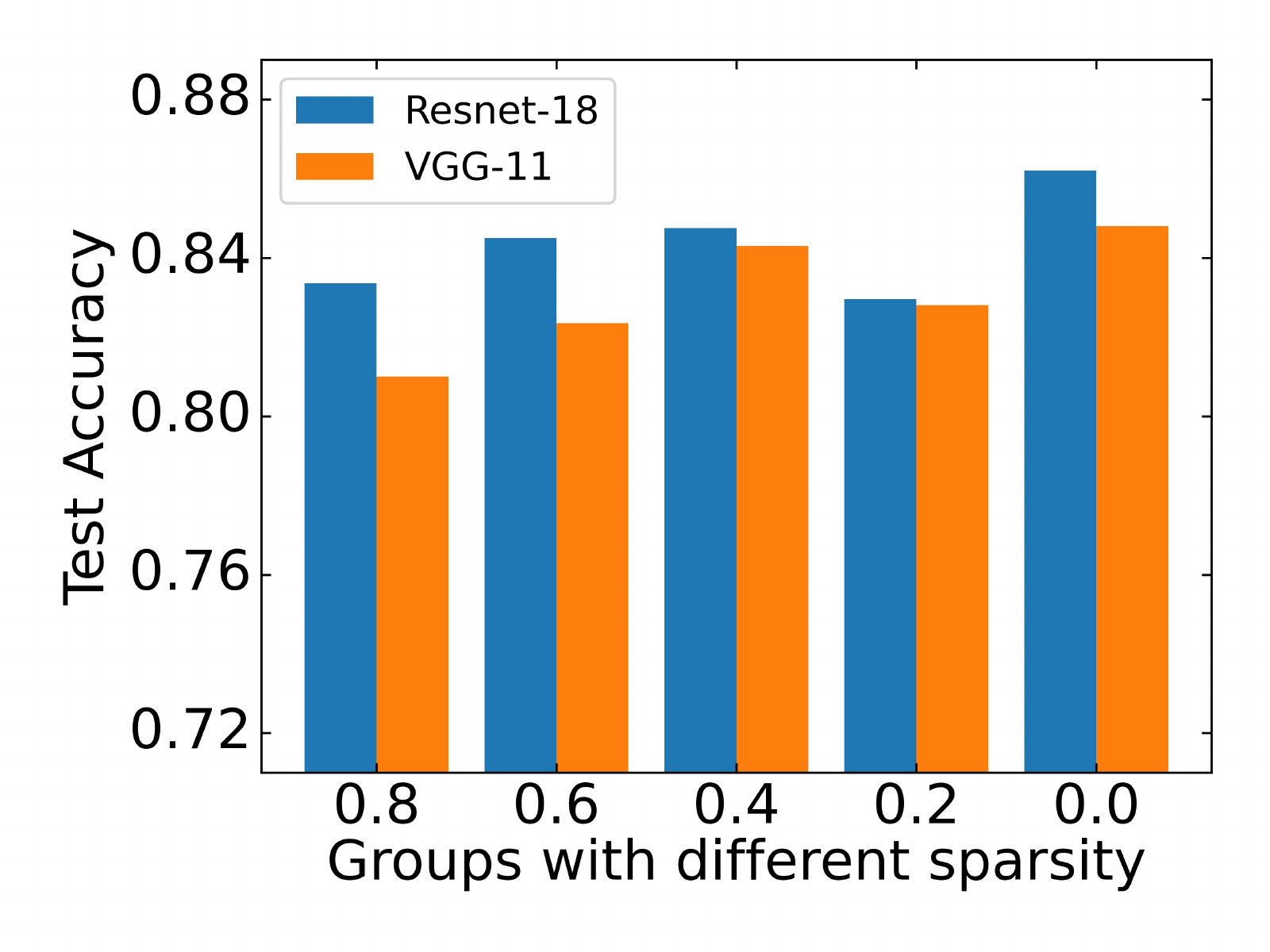}
   \vskip -0.1in
    \subcaption{Dir partition}
  \end{subfigure}
  \hspace{1pt}
  \begin{subfigure}{0.48\linewidth}
   \includegraphics[width=1\linewidth]{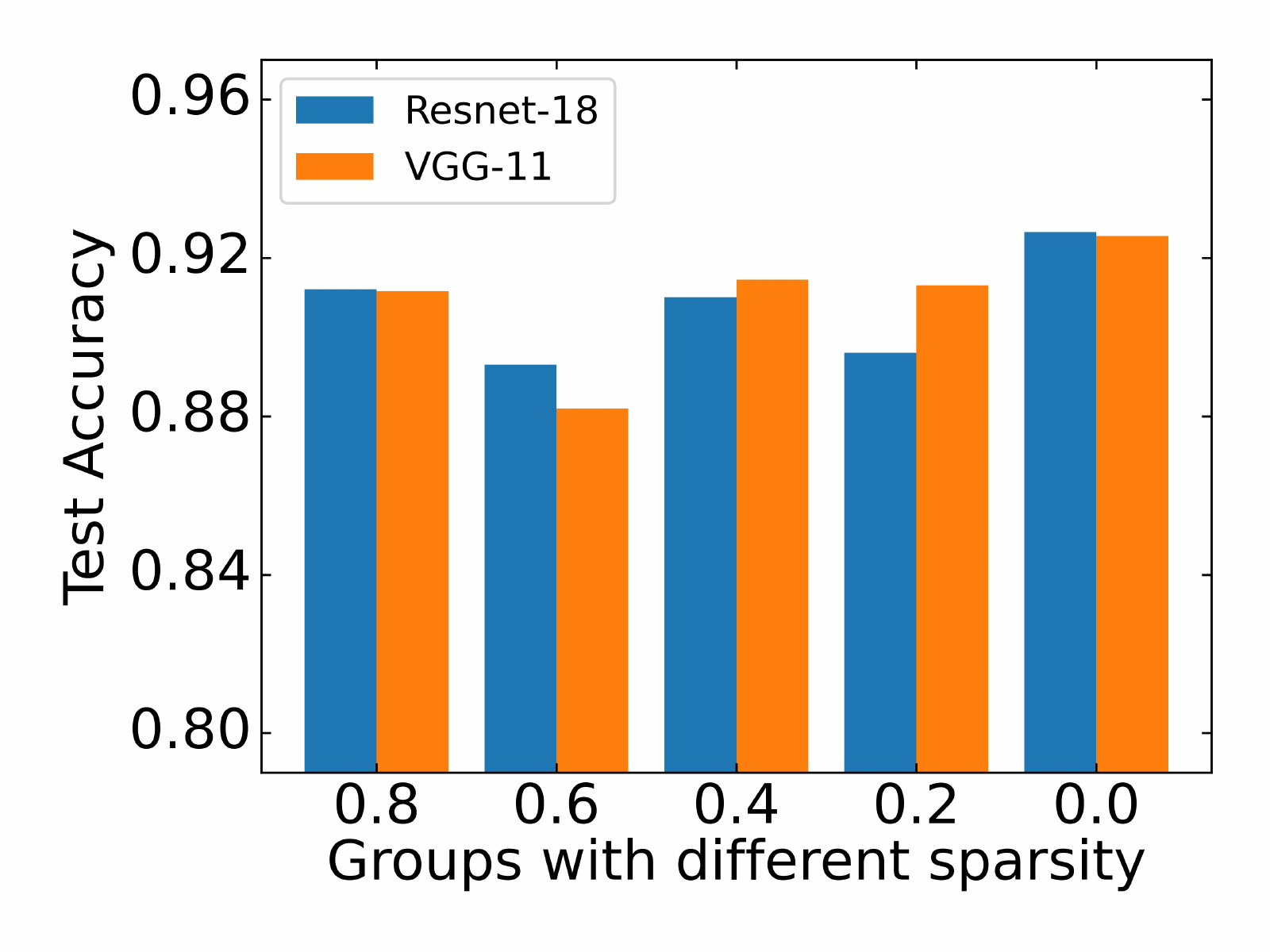}
   \vskip -0.1in
   \subcaption{Path partition}
  \end{subfigure}
 \end{center}
 \vspace{-5pt}
 \caption{Performance of each sparsity group.}
 \label{fig:diff_spa}
 \vskip -0.2in
\end{figure}

\begin{figure}[tb]
 \begin{center}
  \begin{subfigure}{0.49\linewidth}
   \includegraphics[width=1\linewidth]{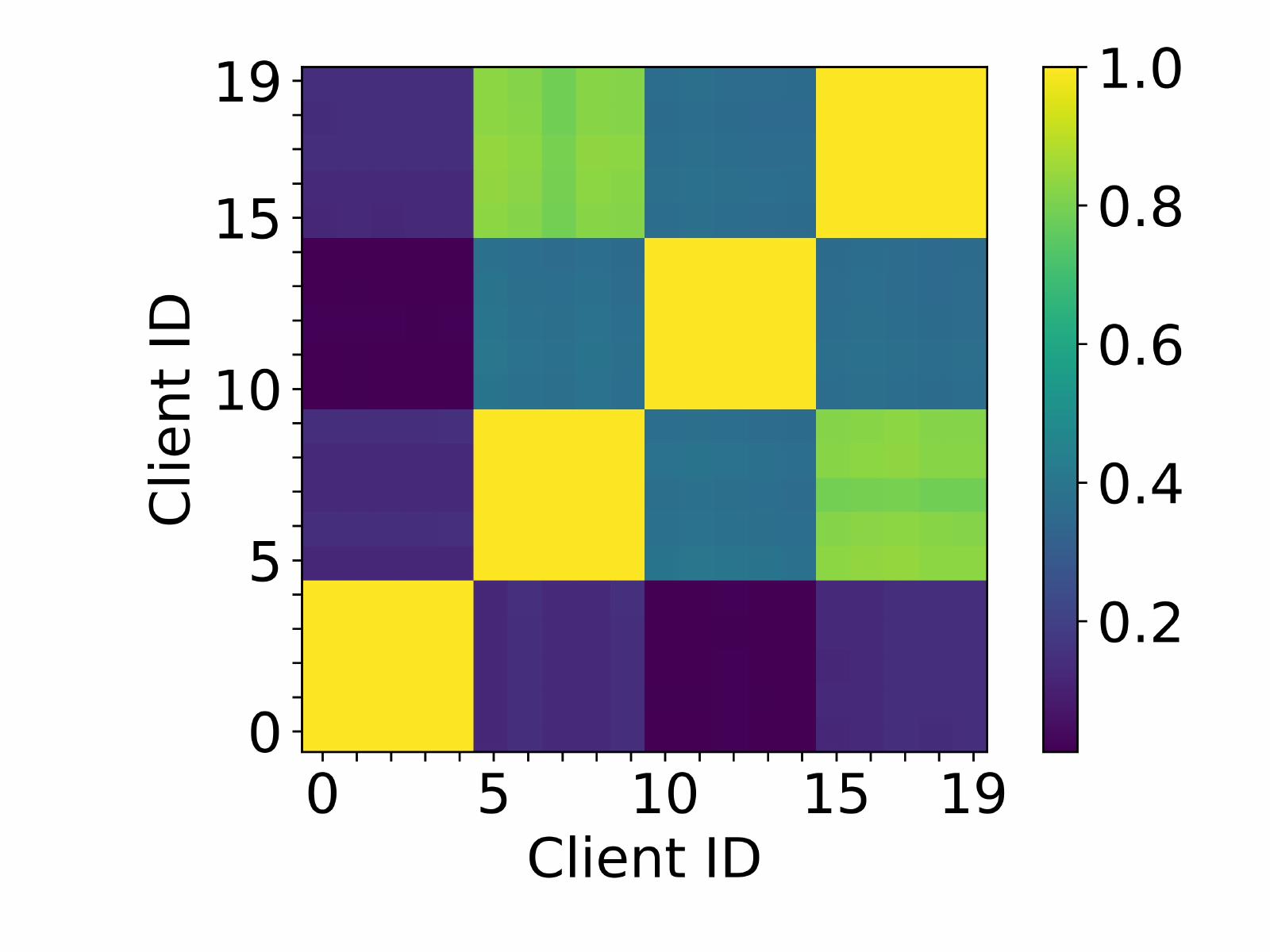}
     \vskip -0.1in
    \subcaption{}
  \end{subfigure}
  \begin{subfigure}{0.49\linewidth}
   \includegraphics[width=1\linewidth]{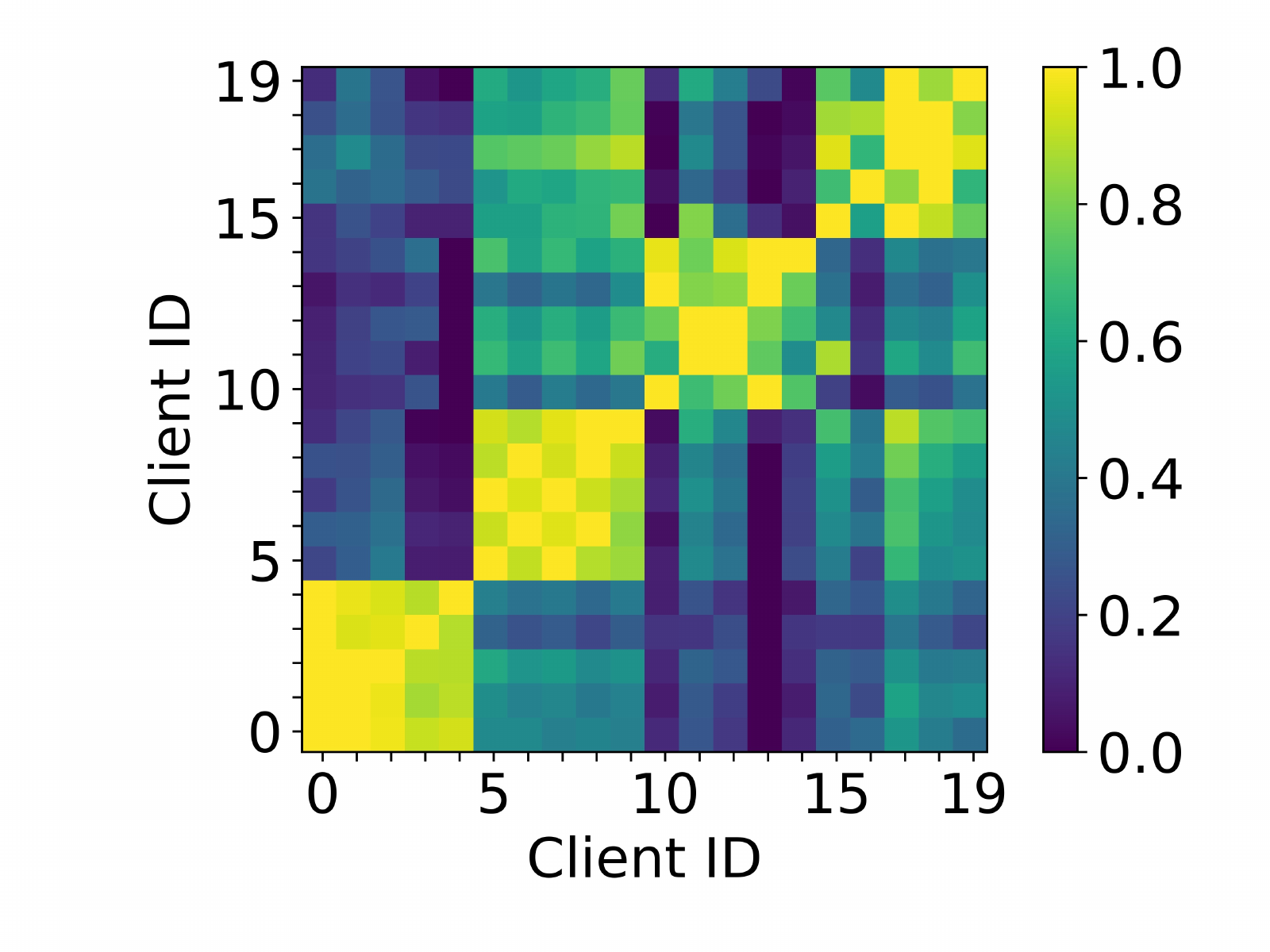}
   \vskip -0.1in
   \subcaption{}
  \end{subfigure}
 \end{center}
 \vspace{-5pt}
 \caption{Explaining the learned masks, while (a) shows the cos-similarity of the training labels distributions between clients and (b) shows the aligned hamming distance between the learned masks.}
 \label{mask-explaining}
 \vskip -0.2in
\end{figure}

\paragraph{Communication cost and local training cost.}
Table \ref{centralized-main-table} and Table \ref{decentralized-main-table} demonstrate that besides achieving better final accuracy, Dis-PFL can also save a lot more peer-to-peer communication costs and reduce the local training flops to a large extent. The edge of our proposed Dis-PFL stems from the training pattern, i.e. training a sparse model through the whole training process. This is also important for the memory constrained setting, dense model may not have the ability to do inference or loss backward pass. Overall, our method can reduce the communication cost and local training cost, which further indicates the inference speedup and the reduced energy consumption.

\paragraph{Convergence speed.}

We illustrate each method's convergence speed via drawing the learning curves of them under different settings as in Figure \ref{fig:main_acc}. We also collect communication rounds for each method to reach a target accuracy in Table \ref{CIFAR10-convergence}-\ref{tiny-convergence} (all in appendix \ref{ap:convergence}). The results show that Dis-PFL requires significantly fewer communication rounds to a target accuracy than other baselines, which indicates a faster convergence. Noticeably, focusing on the communication topology, Table \ref{centralized-main-table} and Table \ref{decentralized-main-table} also demonstrate that with the same communication rounds, the results achieved in the fully connected case may outperform those achieved in the sparsely connected case, e.g ring connected case or dynamic time-varying connected case, indicating a faster convergence. This can be attributed to the increased overall communicated information per round, and it's a natural benefit as shown in \cite{koloskova2019decentralized}. However, it's worth noting that we target at the setting where the busiest node's communication bandwidth is restricted, centralized or decentralized manners are only the communication protocols, our method outperforms other baselines in saving the communication rounds to train the personalized models to a specific accuracy no matter what settings they are.

\subsection{Experiments on client heterogeneous setting}
In the real world, it's common to see that heterogeneous devices take part in one federating process, thus how to deal with the diverse hardware constraints remains a challenging problem. For traditional approaches, the model architecture and size are confined to the weakest device to match each client's hardware constraints. This might have a negative impact on the overall performance since it may not take advantage of today's emerging deeper neural networks. \red{Current works utilize ordered dropout \cite{horvath2021fjord} or splitting technique \cite{diao2020heterofl,hong2022efficient} to best use heterogeneous resources. They all manually design partition methods to part the global model into several parts in order to construct a global model for the union data distribution. However, our method aims at personalization and} can easily adapt to this setting, the overall model architecture can be designed freely, and the sparsity for each client can be chosen wisely according to their own local computation, memory, and communication restrictions.

For demonstration, we run experiments on 100 nodes on CIFAR-10 dataset with Dir partition $\alpha=0.3$ and pathological partition in two settings. (\textbf{i}) All clients have the same capability of computing, saving, and transmitting half of the overall model. (\textbf{ii}) 100 nodes are grouped into 5 parts with the capability of each computing, saving and transmitting 20\%, 40\%, 60\%, 80\% and 100\% parameters of the overall dense model. Since there exists not a central server here, we compare the proposed Dis-PFL with only decentralized FL baselines, D-PSGD and D-PSGD-FT. For setting (\textbf{i}), we implement Dis-PFL with sparsity set 0.5 for all clients, and we implement D-PSGD and D-PSGD-FT with masking half the parameters of the dense model to meet the constraints. For setting (\textbf{ii}), we implement Dis-PFL with assigned correspondingly sparsity in \{0.0, 0.2, 0.4, 0.6, 0.8\} for them. Noticeably, under setting (\textbf{ii}), D-PSGD and D-PSGD-FT can only be implemented with 20\% model size. We choose Resnet-18 and VGG11 as the backbones.

Table \ref{extended-table} shows the results of our methods comparing other baselines in the client heterogeneous setting. It suggests that out approach can easily adapt to this setting. For setting (ii), we can see the performance of D-PSGD and D-PSGD-FT with only 20\% parameters is relatively low due to the fact that they can not embrace the benefit of the newly developing deep models. Another interesting finding is that under a similar communication budget, compared with D-PSGD and D-PSGD-FT implemented with only 50\% parameters, the proposed Dis-PFL can achieve better test accuracy. This can be attributed to the newly designed decentralized sparse training method, which better integrates the information and finds the best suitable mask.

We draw performance of each sparsity group in Figure \ref{fig:diff_spa}. This result show that models with different sparsity, all have the ability to deal with their own tasks in the federating system no matter what the data partition or model backbone they possess. This demonstrates that our method Dis-PFL can well adapted to real life setting where heterogeneous clients may have various types of limits.

\subsection{Empirical analysis of the learned sparse masks. }

In order to explain how the personalized masks produced by our method can help train local personalized models for non-IID tasks, we investigate the correlation between the distance of the learned masks and task similarities. More specifically, we run experiments on a 20 nodes setting on CIFAR-10, where we partition them into 4 groups and each group shares a similar label distribution sampled with Dir(0.3). Task similarities are measured through cos-similarity between two label distributions, while the distances of the learned masks are calculated through the aligned hamming distances. Figure \ref{mask-explaining} demonstrates the correlation between them. It indicates that the personalized masks generated by our methods have the ability to accommodate to the local distribution by not only learning similar masks inside the same group with almost the same latent distributions, but also capturing similar correlations between different groups through assigning corresponding distanced masks. Noticeably, cos-similarities between label distributions and hamming distances between masks both may not best represent the true relationship among different clients. Figure \ref{mask-explaining} is only a simple and straightforward way to demonstrate the relation, deeper insights into the relationship between the learned masks and each client's local distribution are remained for further works.

\subsection{Discussion of the sparsity ratio}


To discover the effects of the sparsity ratio in Dis-PFL, we run experiments on 100 clients under CIFAR-10 Dir Part with different sparsity ratios. Table \ref{tab:spa} demonstrates that it's challenging to find the optimal sparsity ratio. Though a higher sparsity ratio may bring more communication cost benefits, the performance degradation can not be omitted. This can be intuitively understood by the few overlap of the received masks, leading to less information exchange. On the other hand, a small sparsity ratio also performs not well since all clients share almost the same mask, mask personalization technique may not function.

This result further indicates the soundness of Theorem \ref{thm:main}. When the sparsity ratio is small enough (all clients remain an over-parameterized network), as the sparsity ratio grows ($\beta_k$ decreases), the generalization ability of the personalized model grows. However, as the traditional wisdom in machine learning \cite{mohri2018foundations}, there always exists a delicate balance between training error and generalization gap. Though the generalization gap can be controlled when model complexity decreases (greater sparsity), a larger training error may also lead to a bad test error. Thus, in practice, selecting a reasonable sparsity ratio (sweet point) requires significant consideration.

\begin{table}[h]
\small
\caption{Performance of Dis-PFL under different sparsity ratio.}
\label{tab:spa}
\vskip 0.15in
\begin{tabular}{cccccc}
\toprule
Sparsity    & 0.8 & 0.6 & 0.5 & 0.4 & 0.2 \\ \midrule
Acc         & 83.27    & 84.08     & 85.70    & 84.22    & 84.10     \\
Comm (MB)    & 89.30    & 178.69      & 223.4    & 268.1   & 357.5    \\
FLOPS (1e12) & 4.6     & 6.5    & 7.0    & 7.5    & 8.2   \\
\bottomrule
\end{tabular}
\vskip -0.1in
\end{table}

\section{Conclusion}


In this work,we propose Dis-PFL, a {\bf\underline{D}}ecentral{\bf \underline{i}}zed {\bf \underline{s}}parse training based {\bf \underline{P}}ersonnalized {\bf \underline{F}}ederated {\bf \underline{L}}earning approach to simultaneously tackle the data heterogeneity and client heterogeneity for personalized FL. Thanks to the newly designed decentralized sparse training technique, Dis-PFL could reduce the communication bottleneck, save local training costs, and easily adapts to the client heterogeneous setting. Furthermore, we provide theoretical and experimental understandings for the sparse masks. Extensive experiments also verify the efficacy of the proposed Dis-PFL.  

\section*{Acknowledgement}
This work is supported by Science and Technology Innovation 2030 –“Brain Science and Brain-like Research” Major Project (No. 2021ZD0201402 and No. 2021ZD0201405). This work is done during Rong Dai's internship at JD Explore Academy.

\bibliography{main.bib}
\bibliographystyle{icml2022}

\appendix
\onecolumn
\begin{center}
    \bfseries\LARGE Appendix
\end{center}


\section{More details on algorithm implementation}
\subsection{Algorithm 2} \label{algo2}
We present the local mask searching method as in Algorithm 2 here.
\begin{algorithm}[htb]
   \caption{Local mask searching}
   \label{alg:algo2}
\begin{algorithmic}
   \STATE {\bfseries Input:} $\boldsymbol{w}_{k,t+1}$ and corresponding mask $\boldsymbol{m}_{k,t}$
   \STATE {\bfseries Output:} New mask $\boldsymbol{m}_{k,t+1}$
   \STATE Compute current prune rate $\alpha_t$ using cosine annealing principle with initial pruning rate $\alpha_0$
   \STATE Sample a batch of local data and do loss backward to get the dense gradient $g({w}_{k,t+1})$
   \FOR {layer $j$ $\in$ $J$ do}
   \STATE Update mask $m^{j}_{k,t+\frac{1}{2}}$ by zeroing out $\alpha_t$-proportion of weights with magnitude pruning
   \STATE Update mask $m^{j}_{k,t+1}$ via recovering weights with gradient information $g({w}_{k,t+1})$
   \ENDFOR
   \STATE Get new mask $m_{k,t+1}$

\end{algorithmic}
\end{algorithm}

\section{Experiments}\label{ap:exp}
In this section, we provide more details of our experiments and more extensive experimental results to compare the performance of the proposed Dis-PFL against other baselines.
\subsection{Datasets}
We use CIFAR-10 (10 classes, 5000 training samples each), CIFAR-100 (100 classes, 500 training samples each), and Tiny-Imagenet (200 classes, 500 training samples each) for the experiments. We use two non-IID partition methods to split the training data in our implementation. One is based on the Dirichlet distribution on the label ratios to ensure uneven label distributions among devices as in \cite{hsu2019measuring}, a smaller $\alpha$ indicates higher data heterogeneity. The other is called pathological partition, which means only assigning samples of specific classes for each client. To simulate the personalized FL setting, each client's testing data has the same proportion of labels as its training data, and the total number of the testing set is set to 100 for all partitions.

\subsection{Model Architectures}
We follow the pytorch's implementation of ResNet18 \cite{he2016deep} and VGG11 \cite{vgg11} to do all the evaluations. Since batch-norm may have a detrimental effect on federated learning \cite{hsieh2020non}, we replace all the batch-norm layers in ResNet18 and VGG11 with group-norm layers \cite{wu2018group}. 

\subsection{Hyper-Parameters}
We use SGD optimizer for all methods with weighted decayed parameter 0.0005. For all the methods except \textbf{Ditto}, local epochs are fixed to 5. For \textbf{Ditto}, in order to ensure a fair comparison, each client performs 3 epochs for training the local model and 2 epochs for training the global model. The learning rate is initialized with 0.1 and decayed with 0.998 after each communication round. The batch size is fixed to 128 for all the experiments. We run 500 global communication rounds for CIFAR-10, CIFAR-100, and 300 for Tiny-Imagenet. The setting of hype-parameters also impacts the generalizability \cite{he2019control}.

\subsection{More details about baselines}
\textbf{Local} is the direct solution to the personalized federated learning problem. Each client only performs SGD on their own data. For the sake of consistency, we take 5 epochs of local training as one communication round. \textbf{FedAvg \cite{mcmahan2017communication}} is the most widely studied FL method. The vanilla weighted average is used to enable all the clients to collaboratively train a global model. \textbf{FedAvg-FT}\cite{cheng2021fine} is a simple method by doing some fine-tuning steps with local data after acquiring the global model but shown to be competitive against various personalized federated learning specific methods. \textbf{Ditto} \cite{li2021ditto} achieves personalization via trade-off between the global model and local objectives. Specifically, within each communication round, each client first trains the global model (similarly aggregated as in FedAvg) on its local empirical risk. And then additionally trains its local model based on a loss function combining its local empirical loss and the proximal term towards the global model. \textbf{FOMO} \cite{zhang2020personalized} trains personalized models using neighbors' gradient information to infer how much a client can benefit from another's model and thus get the adaptive mixing weights for personalization. \textbf{SubFedAvg} \cite{vahidian2021personalized} maintains personalized sub-networks for each user. The overall training process follows a dense-to-spare training rule, the client's local model starts from a fully dense model, and is iteratively pruned as the training progresses. \textbf{D-PSGD} \cite{lian2017can} is a classic decentralized parallel stochastic gradient descent method proposed to reach a consensus model on a decentralized network. Each node first averages the local variables with the received models and then updates it using the local stochastic gradient. To extend it to the federated learning scenarios, following \cite{sun2021decentralized}, we take several epochs of local training instead of one iteration for each local client. We also extend it to \textbf{D-PSGD-FT} following the idea of \textbf{FedAvg-FT}, several steps of local fine-tuning are done to further personalize the global consensus model towards heterogeneous clients.

\subsection{More experiments results}
\subsubsection{Convergence speed} \label{ap:convergence}
We demonstrate the needed communication rounds for each method to reach a target accuracy as follows. Results for CIFAR-10 are in Table \ref{CIFAR10-convergence}, while results for CIFAR-100 in Table \ref{CIFAR100-convergence} and Tiny-Imagenet in Table \ref{tiny-convergence}.

\begin{table}[htb]
\centering
\small
\caption{Averaged needed communication rounds for each method to a target accuracy on CIFAR-10 dataset.}
\label{CIFAR10-convergence}
\vskip 0.15in
\begin{tabular}{cccc|ccc}
\toprule
\multirow{2}{*}{Methods} & \multicolumn{3}{c|}{Dir partition} & \multicolumn{3}{c}{Pathological partition} \\
                         & Acc@60      & Acc@70      & Acc@80      & Acc@50         & Acc@80         & Acc@85         \\ \midrule
FedAvg  & 159          & 267          & \textgreater 500         
& 388   & \textgreater 500  & \textgreater 500      \\
D-PSGD  & 139          & 236          & \textgreater 500           
& 391   & \textgreater 500 & \textgreater 500        \\
Ditto   & 176          & 375          & \textgreater 500           
& 21  & 138     & 256             \\
FOMO & 204 & \textgreater 500 & \textgreater 500  
& 3 & 45     & 129             \\
SubFedAvg  & 67  & \textbf{115}  & \textgreater 500 
& 33  & 99  & 181              \\
Dis-PFL & \textbf{59}  & 144 & \textbf{301}          
& \textbf{3} & \textbf{33} & \textbf{81}   \\ \bottomrule
\end{tabular}
\vskip -0.1in
\end{table}

\begin{table*}[htb]
\centering
\caption{Averaged needed communication rounds for each method to a target accuracy on CIFAR-100 dataset.}
\small
\label{CIFAR100-convergence}
\vskip 0.15in
\begin{tabular}{cccc|ccc}
\toprule
\multirow{2}{*}{Methods} & \multicolumn{3}{c|}{Dir partition} & \multicolumn{3}{c}{Pathological partition} \\
                         & Acc@25      & Acc@40      & Acc@50      & Acc@30         & Acc@50         & Acc@60         \\ \midrule
FedAvg  & 195          & 454          & \textgreater 500         
& 393   & \textgreater 500  & \textgreater 500      \\
D-PSGD  & 157          & 437          & \textgreater 500           
& 326   & \textgreater 500 & \textgreater 500        \\
Ditto   & 201          & \textgreater 500 & \textgreater 500           
& 136  & 393     & \textgreater 500             \\
FOMO & 171 & \textgreater 500 & \textgreater 500  
& \textbf{23} & 223     & \textgreater 500             \\
SubFedAvg  & 161  & \textbf{228}  & \textgreater 500 
& 148  & 202  & 377              \\
Dis-PFL & \textbf{64}  & 231 & \textbf{393}          
& 29 & \textbf{159} & \textbf{293}   \\ \bottomrule
\end{tabular}
\vskip -0.1in
\end{table*}

\begin{table*}[htb]
\centering
\caption{Averaged needed communication rounds for each method to a target accuracy on Tiny-Imagenet dataset.}
\small
\label{tiny-convergence}
\vskip 0.15in
\begin{tabular}{cccc|ccc}
\toprule
\multirow{2}{*}{Methods} & \multicolumn{3}{c|}{Dir partition} & \multicolumn{3}{c}{Pathological partition} \\
                         & Acc@05      & Acc@10      & Acc@15      & Acc@05         & Acc@10         & Acc@20         \\ \midrule
FedAvg  & 92          & 206          & \textgreater 300        
& 89   & \textgreater 300  & \textgreater 300      \\
D-PSGD  & 61   & 141   & \textgreater 300           
& 66   & 185 & \textgreater 300        \\
Ditto   & 39 & 137 & 261           
& 19  & 84    & 211             \\
FOMO & 168 & \textgreater 300 & \textgreater 300  
& \textbf{2} & \textgreater 300  & \textgreater 300             \\
SubFedAvg  &  80 & 201  & \textgreater 300 
& 88  & 222  & \textgreater 300       \\
Dis-PFL & \textbf{13}  & \textbf{58} & \textbf{195}          
& 3 & \textbf{18} & \textbf{63}   \\ \bottomrule
\end{tabular}
\vskip -0.1in
\end{table*} 

\newpage

\subsubsection{Different topology} \label{ap:top}
Similar to Table \ref{decentralized-main-table}, we also record performances of each method under different decentralized topology in Table \ref{tab:de-cifar100} for CIFAR-100 and Table \ref{tab:de-tiny} for Tiny-Imagenet.

\begin{table*}[tbh]
\caption{Table illustrating performance compared with methods in decentralized communication protocols on CIFAR-100.}
\label{tab:de-cifar100}
\vskip 0.1in
\centering
\small
\begin{tabular}{ccccccc}
\toprule
\multirow{2}{*}{Dataset}       & \multirow{2}{*}{Topology}        & \multirow{2}{*}{Method} & \multirow{2}{*}{\begin{tabular}[c]{@{}c@{}}Dir Part\\ Acc\end{tabular}} & \multirow{2}{*}{\begin{tabular}[c]{@{}c@{}}Path Part\\ Acc\end{tabular}} & \multirow{2}{*}{\begin{tabular}[c]{@{}c@{}}Comm\\(MB)\end{tabular}} & \multirow{2}{*}{\begin{tabular}[c]{@{}c@{}}FLOPS\\(1e12)\end{tabular}} \\
                               &                            &                         &                                                                    &                                                                     &                                                                                   &                                                                                      \\
                              \midrule
\multirow{7}{*}{CIFAR-100}     & Seperate                         & Local                   & 29.23$\pm$0.2                                                                   & 52.46$\pm$0.2                                                               & -                                                                                  & 8.3                                                                                     \\
\cmidrule{2-7}
                               & \multirow{3}{*}{Ring}            & D-PSGD                  & 17.52$\pm$0.2                                                                   & 10.62$\pm$0.5                                                               & 89.7                                                                                  & 8.3                                                                                     \\
                               &                                  & D-PSGD-FT               & 32.61$\pm$0.3                                                                   & 51.67$\pm$0.2                                                               & 89.7                                                                                  & 8.3                                                                                     \\
                               &                                  & Dis-PFL(ours)           & \textbf{33.13$\pm$0.2}                                                                   & \textbf{52.08$\pm$0.3}                                                      & \textbf{44.8}                                                                                 &                                                                                      \\
                               \cmidrule{2-7}
                               & \multirow{3}{*}{Fully-connected} & D-PSGD                  & 43.18$\pm$0.4                                                              & 36.81$\pm$0.4                                                               & 4442.2                                                                                  & 8.3                                                                                     \\
                               &                                  & D-PSGD-FT               & \textbf{53.40$\pm$0.2}                                                              & 68.23$\pm$0.2                                                              & 4442.2                                                                                   & 8.3                                                                                     \\
                               &                                  & Dis-PFL(ours)           & 52.85$\pm$0.2                                                             & \textbf{72.97$\pm$0.5}                                                      & \textbf{2222.0}                                                                                  & \textbf{7.0}                                                                                     \\

                               \bottomrule
\end{tabular}

\end{table*}

\begin{table*}[htb]
\caption{Table illustrating performance compared with methods in decentralized communication protocols on Tiny-Imagenet.}
\label{tab:de-tiny}
\vskip 0.1in
\centering
\small
\begin{tabular}{ccccccc}
\toprule
\multirow{2}{*}{Dataset}       & \multirow{2}{*}{Topology}        & \multirow{2}{*}{Method} & \multirow{2}{*}{\begin{tabular}[c]{@{}c@{}}Dir Part\\ Acc\end{tabular}} & \multirow{2}{*}{\begin{tabular}[c]{@{}c@{}}Path Part\\ Acc\end{tabular}} & \multirow{2}{*}{\begin{tabular}[c]{@{}c@{}}Comm\\(MB)\end{tabular}} & \multirow{2}{*}{\begin{tabular}[c]{@{}c@{}}FLOPS\\(1e12)\end{tabular}} \\
                               &                            &                         &                                                                    &                                                                     &                                                                                   &                                                                                      \\
                              \midrule
\multirow{7}{*}{Tiny-Imagenet} & Seperate                         & Local                   & 6.76$\pm$0.2                                                                   & 17.68$\pm$0.3                                                                    & -                                                                                  & 6.7                                                                                     \\
\cmidrule{2-7}
                               & \multirow{3}{*}{Ring}            & D-PSGD                  & 3.63$\pm$0.4                                                                   & 3.02$\pm$0.5                                                                    & 90.1                                                                                  & 66.6                                                                                     \\
                               &                                  & D-PSGD-FT               & \textbf{9.66$\pm$0.2}                                                                  & 19.72$\pm$0.4                                                                   & 90.1                                                                                  & 66.6                                                                                     \\
                               &                                  & Dis-PFL(ours)           & 9.60$\pm$0.2                                                                   & \textbf{20.17$\pm$0.2}                                                                    & \textbf{45.0}                                                                                  & \textbf{54.5}                                                                                      \\
                               \cmidrule{2-7}
                               & \multirow{3}{*}{Fully-connected} & D-PSGD                  & 12.92$\pm$0.3                                                                   & 11.58$\pm$0.3                                                                    & 4462.5                                                                                  & 66.6                                                                                     \\
                               &                                  & D-PSGD-FT               & 16.52$\pm$0.2                                                                    & 29.76$\pm$0.3                                                                    & 4462.5                                                                                  & 66.6                                                                                     \\
                               &                                  & Dis-PFL(ours)           & \textbf{17.10$\pm$0.3}                                                                  & \textbf{31.93$\pm$0.2}                                                                   & \textbf{2229.8}                                                                                  & \textbf{54.5}
                               \\
                               \bottomrule
\end{tabular}

\end{table*}

\subsection{Extended experiments on random clients dropping settings} \label{ap:drop}
It's common to see that clients or the server may fail to all take part in every communication round in the federated system. A server malfunction may hurt the overall system, thus leading to the failed operation of the whole system. However, this weakness can be alleviated in the decentralized setting, since one or more clients drop this round may not hurt the overall training process. We here conduct the dropped clients experiments on Dis-PFL to demonstrate the robustness to random client dropping of the proposed method. The experiments are conducted on CIFAR-10 with 100 clients under Dir partition.
\begin{figure}[htb]
 \begin{center}
   \includegraphics[width=0.6\linewidth]{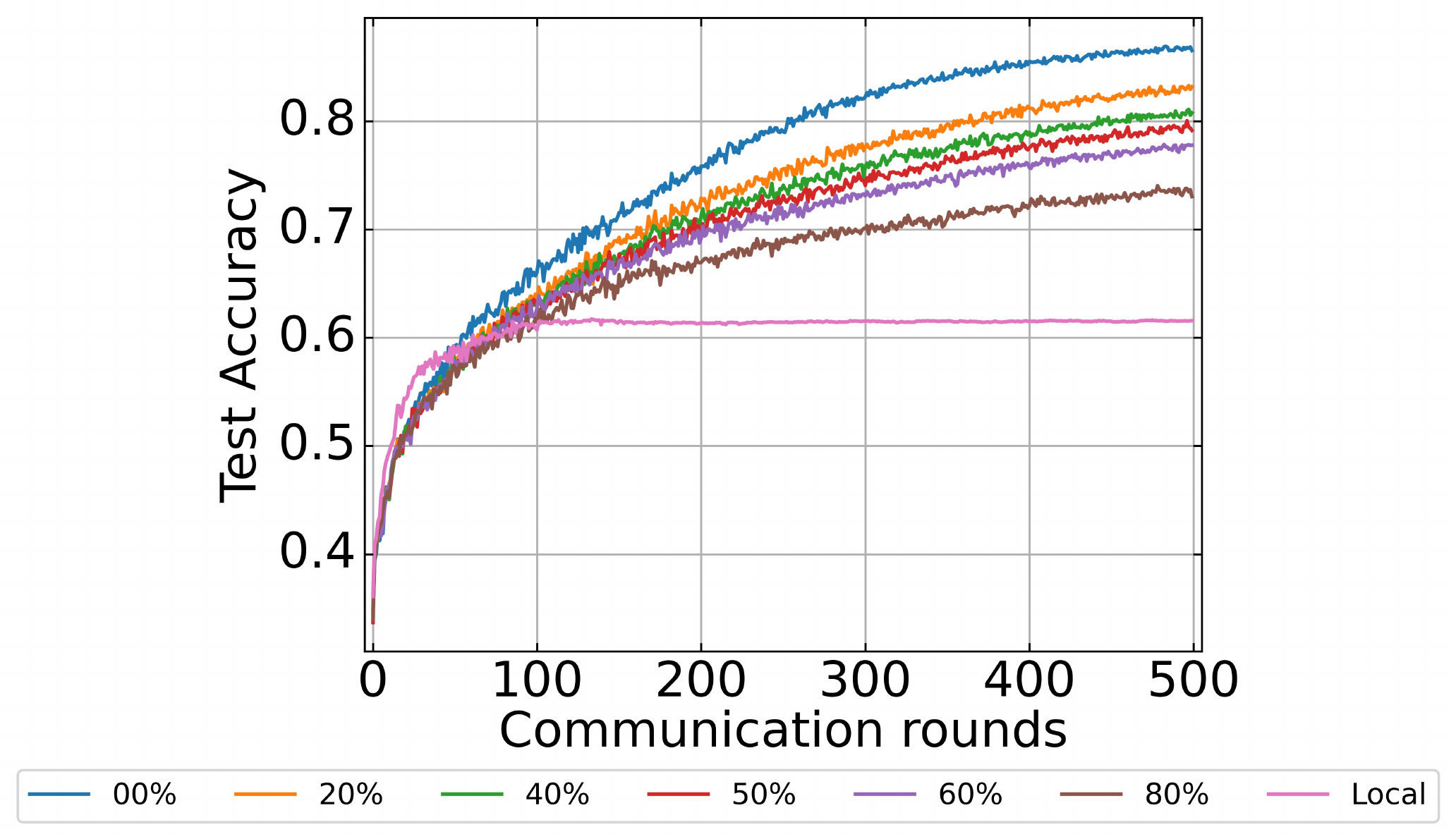}
 \end{center}
 \vspace{-5pt}
 \caption{Performance of Dis-PFL under different clients dropping probabilities in the fully connected topology}
 \label{fig:drop}
 \vskip -0.1in
\end{figure}

As shown in Figure \ref{fig:drop}, our proposed Dis-PFL can provide fairly decent personalized models for each device participating in the federation system compared with local training, regardless of the device dropping probability. Admittedly, as the probability of device disconnection increases, the convergence speed of the Dis-PFL may slow down, and it may also affect the final personalized result. Nevertheless, our proposed decentralized personalization algorithm is still more robust than the centralized personalization algorithm since if the server drops in a certain communication round, the entire centralized system cannot work at all.

\section{Proof of Theorem \ref{thm:main}} \label{ap:proof}
This section demonstrates the proof of Theorem \ref{thm:main} in detail. We first clarify the notations and preliminaries in \ref{ap:np} and present some key lemmas in \ref{ap:kl} and finally present the proof in \ref{ap:mp}.

\subsection{Notations and Preliminaries} \label{ap:np}
$S = \{(x_1, y_1), \ldots, (x_N, y_N) | x_i \in \mathcal X \subset \mathbb R^{d_X},$ $ y_i \in \mathcal Y \subset \mathbb R^{d_Y}, i = 1, \ldots, N\}$ is a training sample set, where $x_i$ is the $i$-th feature, $y_i$ is the corresponding label, and $d_X$ and $d_Y$ are the dimensions of the feature and the label, respectively. For the brevity, we define $z_i = (x_i, y_i)$. We also define random variables $Z = (X, Y)$, such that all $z_i = (x_i, y_i)$ are independent and identically distributed (i.i.d.) observations of the variable $Z = (X, Y) \in \mathcal Z,~ Z \sim \mathcal D$, where $\mathcal D$ is the data distribution.

For a machine learning algorithm $\mathcal A$, it learns a hypothesis $\mathcal A(S)$, $\mathcal A(S) \in \mathcal H \subset \mathcal Y^{\mathcal X} = \{f: \mathcal X \to \mathcal Y\}$. 

The expected risk $\mathcal{R}_{D}(\mathcal A(S))$ and empirical risk $\hat{\mathcal R}_{\mathcal{S}}(\mathcal{A}(S))$ of the algorithm $\mathcal{A}$ are defined as follows,
\begin{gather*}
	 \mathcal{R}_{\mathcal{D}}(\mathcal{A}(S)) = \mathbb{E}_{z\sim \mathcal{D}} \ell (\mathcal{A}(S), z),\\
	\hat{\mathcal R}_S(\mathcal{A}(S)) = \frac{1}{N} \sum_{i=1}^N \ell(\mathcal{A}(S), z_i),
\end{gather*}
where $\ell: \mathcal H \times \mathcal Z \to \mathbb R^+$ is the loss function. 

\begin{definition}[Generalization bound]
\label{ap:gb}
The generalization error is defined as the difference between the expected risk and empirical risk, 
\begin{equation*}
\operatorname{Gen}_{S,\mathcal{A}(S)}\overset{\triangle}{=} \mathcal{R}_{\mathcal{D}}(\mathcal{A}(S)) - \hat{\mathcal R}_S (\mathcal{A}(S)),
\end{equation*}
whose upper bound is called the generalization bound.

\end{definition}

\begin{definition}[Differential Privacy]
\label{ap:dp}
A stochastic algorithm $\mathcal{A}$ is called ($\varepsilon,\delta$)-differentially private if for any hypothesis subset $\mathcal{H}_0 \subset \mathcal H$ and any neighboring sample set pair $S$ and $S'$ which differ by only one example (called $S$ and $S'$ adjacent), we have
\begin{equation*}
\label{eq:dp}
\log \left[ \frac{\mathbb P_{\mathcal{A}(S)}(\mathcal{A}(S)\in \mathcal{H}_0) - \delta}{\mathbb P_{\mathcal{A}(S')}(\mathcal{A}(S')\in \mathcal{H}_0)} \right] \le \varepsilon.
\end{equation*}
The algorithm $\mathcal{A}$ is also called $\varepsilon$-differentially private, if it is $(\varepsilon, 0)$-differentially private.
\end{definition}

\begin{definition}[Multi-Sample-Set Learning Algorithms]
Suppose the training sample set $S$ with size $kN$ is separated to $k$ sub-sample-sets $S_1, \ldots, S_k$, each of which has the size of $N$. In another word, $S$ is formed by $k$ sub-sample-sets as
\begin{equation*}
S = (S_1, \ldots, S_k).
\end{equation*}
The hypothesis $\mathcal{B}(S)$ learned by {\it multi-sample-set algorithm} $\mathcal{B}$ on dataset $S$ is defined as follows,
\begin{equation*}
\mathcal{B}: \mathcal Z^{k\times N}\mapsto \mathcal H \times \{1, \ldots, k\},~ \mathcal{B}(S)=  \left(h_{\mathcal B(S)}, i_{\mathcal B(S)} \right).
\end{equation*}
\end{definition}

\subsection{Key Lemmas}\label{ap:kl}

\begin{lemma}[c.f. Theorem 9 \citet{balle2018privacy}] 
\label{lemma:one}
This lemma provide bound of differential privacy parameters after sub-sampling uniformly without replacement. Let $\mathcal{M}^o:\mathcal{Z}^m\mapsto \Delta \mathcal{H}$ be any mechanism preserving $(\varepsilon,\delta)$ differential privacy. Let   $\mathcal{M}^{wo}:\mathcal{Z}^N\mapsto \Delta \mathcal{Z}^m$ be the uniform sub-sampling without replacement mechanism. Then mechanism $\mathcal{M}^o\circ\mathcal{M}^{wo}$ satisfy $(\log (1+(m / N)(e^{\varepsilon}-1)),m\delta/N)$  differential privacy.
\end{lemma}

\begin{lemma}[c.f. Theorem 4 \citet{he2021tighter}] 
\label{lemma:multi}
This lemma gives the relationship between one step privacy preserving methods and iterative machine learning methods. Suppose an iterative machine learning algorithm $\mathcal A$ has $T$ steps: $\left\{W_i(S)\right\}_{i=1}^T$. Specifically, we define the $i$-th iterator as follows,
	\begin{equation*}
	\mathcal{M}_i: (W_{i-1}(S), S) \mapsto W_{i}(S).
	\end{equation*}
	Assume that $W_0$ is the initial hypothesis (which does not depend on $S$). If for any fixed $W_{i-1}$, $\mathcal{M}_i(W_{i-1},S)$ is $\varepsilon_i$-differentially private, 
	then $\left\{W_i\right\}_{i=0}^T$ is ($\varepsilon'$, $\delta'$)-differentially private that
	\begin{equation*}
	\varepsilon^{\prime}=\sqrt{2  \log \left( \frac{1}{\delta'}\right)\left(\sum\limits_{i=1}^T \varepsilon_{i}^2\right)} +\sum\limits_{i=1}^T \varepsilon_i \frac{e^{\varepsilon_i}-1}{e^{\varepsilon_i}+1},
	\end{equation*}
	\begin{align*}
    \delta' = & e^{-\frac{\varepsilon'+T\varepsilon}{2}}\left(\frac{1}{1+e^\varepsilon}\left(\frac{2T\varepsilon}{T\varepsilon-\varepsilon'}\right)\right)^T\left(\frac{T\varepsilon+\varepsilon'}{T\varepsilon-\varepsilon'}\right)^{-\frac{\varepsilon'+T\varepsilon}{2\varepsilon}}- \left(1-\frac{\delta}{1+e^{\varepsilon}}\right)^{T}
\\
&+2-\left(1-e^{\varepsilon}\frac{\delta}{1+e^{\varepsilon}}\right)^{\left \lceil  \frac{\varepsilon'}{\varepsilon}\right \rceil}\left(1-\frac{\delta}{1+e^{\varepsilon}}\right)^{T-\left \lceil  \frac{\varepsilon'}{\varepsilon}\right \rceil} .
\end{align*}
\end{lemma}

\begin{lemma}[c.f. Theorem 1 \citet{he2021tighter}] 
\label{lemma:g-dp}
This lemma gives a high-probability generalization bound for any ($\varepsilon,\delta$)-differentially private machine learning algorithm. Suppose algorithm $\mathcal{A}$ is ($\varepsilon,\delta$)-differentially private, the training sample size $N\ge\frac{2}{\varepsilon^{2}} \ln \left(\frac{16}{e^{-\varepsilon}\delta}\right)$, and the loss function $\Vert l\Vert_{\infty}\le 1$. Then, for any data distribution $\mathcal{D}$ over data space $\mathcal{Z}$, we have the following inequality,
\begin{equation*}
\mathbb{P}\left[\left|\hat{\mathcal{R}}_S(\mathcal{A}(S)) - \mathcal{R}(\mathcal{A}(S))\right| < 9\varepsilon\right] > 1-\frac{e^{-\varepsilon}\delta}{\varepsilon} \ln \left(\frac{2}{\varepsilon}\right).
\end{equation*}
\end{lemma}

\subsection{Main proof} \label{ap:mp}

\begin{proof}
The main proof can be seen as acquiring generalization bound through the lens of differential privacy. The proof skeleton can be concluded in three stages: (1) We first take a global view of the proposed algorithm Dis-PFL and thus classify it as an iterative machine learning algorithm. (2) We then calculate the differential privacy of each step in the algorithm. (3) Extend it to iterative situations and acquire the final result through the bridges provided in \cite{he2021tighter}.

First, let us take a global view of the overall decentralized training process. We can assume that the initial consensus model derived using the model fusion approach in the newly designed decentralized sparse training technique is the same for all clients. Afterward, each client multiplies the personalized mask with this consensus model to get the specific initial model for them to further operate local sparse training and mask searching. Thus, a global consensus model $\boldsymbol{w}$ always exists during the overall training process. The proportion of remaining parameters for this model $\beta$ is inferred from the aggregation of all the local clients with $\beta_k$ remaining params. 

For simplicity, we define $W_t$ as the virtual consensus model at iteration $t$. Then the decentralized learning paradigm can be seen as iteratively optimize $W$ using partial gradient information (due to the sparse mask) on each client $k$. We also denote $\mathcal{N}(0, \sigma^2\mathbb{I})$ as a Gaussian noise, where $\sigma$ is the Gaussian noise variance. We define $\tau$ as the mini-batch size and overall iteration steps as $T$. We assume the computed gradient of the loss function $\mathcal{L}$ is bounded, and the diameter of the gradient space is defined as $D_g\overset{\triangle}{=}\max_{W,z,z'}\Vert \nabla\ell(z,W)- \nabla\ell(z',W)\Vert$. We also denote $G_{\mathcal{B}}(W)\overset{\triangle}{=}\frac{1}{\Vert \mathcal{B}\Vert}\sum_{z\in \mathcal{B}}g(z,W)$ as the mean of $g$ over $\mathcal{B}$ for brevity. We also use $\boldsymbol{p}$ as the probability density, with $\boldsymbol{p}^{V}$ the probability density conditional on any random variable $V$.

Then we calculate the differential privacy of each step. Recall Algorithm \ref{alg:algo1}, line 9-13 denotes the local training process, the gradient information, $\bigtriangledown_{\widetilde{\boldsymbol{w}}_{k,t,\tau}}$ is calculated on the subset of local data. However, in real federating system, for privacy concerns, additive Gaussian noise sample is also used to enhance privacy. Line 10 in Algorithm \ref{alg:algo1} is equivalent to uniformly sampling a mini-batch $\mathcal{I}_t$ from index set $[N]$ with size $\tau$ without replacement and letting $\mathcal{B}_t=S_{\mathcal{I}_t}$. Furthermore, for fixed $W_{t-1}$, $\mathcal{I}$, and any two adjacent sample sets $S$ and $S'$, we have
\begin{equation}
	\begin{aligned}
    \frac{\boldsymbol{p}^{S,\mathcal{I}_t}(W_{t}=W\vert W_{t-1})}{\boldsymbol{p}^{S',\mathcal{I}_t}(W_{t}=W\vert W_{t-1})}
	&=\frac{\boldsymbol{p}^{S,\mathcal{I}_t}(\eta_t(G_{S_{\mathcal{I}}}(W_{t-1})+\mathcal{N}(0, \sigma^2\mathbb{I}))=W-W_{t-1})}{\boldsymbol{p}^{S',\mathcal{I}_t}(\eta_t(G_{S'_{\mathcal{I}}}(W_{t-1})+\mathcal{N}(0, \sigma^2\mathbb{I}))=W-W_{t-1})}
	\\
	&=\frac{\boldsymbol{p}^{\mathcal{I}_t,W_{t-1}}(\mathcal{N}(0, \sigma^2\mathbb{I})=W')}{\boldsymbol{p}^{S,S',\mathcal{I}_t,W_{t-1}}(G_{S'_{\mathcal{I}}}(W_{t-1})-G_{S_{\mathcal{I}}}(W_{t-1})+\mathcal{N}(0, \sigma^2\mathbb{I})=W')},
	\end{aligned}
\end{equation}
where $\eta_t W'=W-W_{t-1}-\eta_t G_{S_{\mathcal{I}}}(W_{t-1})$. Therefore, when consider the additive Gaussian noise into consideration, if $W\sim W_{t-1}+\eta_t (G_{S_{\mathcal{I}}}(W_{t-1})+\mathcal{N}(0,\sigma \mathbb{I}))$, then $W'\sim G_{S_{\mathcal{I}}}(W_{t-1})+\mathcal{N}(0,\sigma \mathbb{I}) $.

For simplicity, according to the definition of differential privacy, we define 
	\begin{equation}
	D_p^{S,S',\mathcal{I}_t,W_{t-1}}(W')=\log \frac{\boldsymbol{p}^{\mathcal{I}_t,W_{t-1}}(\mathcal{N}(0, \sigma^2\mathbb{I})=W')}{\boldsymbol{p}^{S,S',\mathcal{I}_t,W_{t-1}}(G_{S'_{\mathcal{I}}}(W_{t-1})-G_{S_{\mathcal{I}}}(W_{t-1})+\mathcal{N}(0, \sigma^2\mathbb{I})=W')},
	\end{equation}
which by the definition of Gaussian distribution further leads to
\begin{equation}
    \begin{aligned}
	D_p(W')
	=&-\frac{\Vert W'\Vert^2}{2\sigma^2}+\frac{\Vert W'-G_{S'_{\mathcal{I}}}(W_{t-1})+G_{S_{\mathcal{I}}}(W_{t-1})\Vert^2}{2\sigma^2}
	\\
	=&\frac{2\langle W',-G_{S'_{\mathcal{I}}}(W_{t-1})+G_{S_{\mathcal{I}}}(W_{t-1})\rangle+\Vert G_{S'_{\mathcal{I}}}(W_{t-1})-G_{S_{\mathcal{I}}}(W_{t-1})\Vert^2}{2\sigma^2}.
	\end{aligned}
\end{equation}

Denote $-G_{S'_{\mathcal{I}}}(W_{t-1})+G_{S_{\mathcal{I}}}(W_{t-1})$ as $\mathbf{v}$. By the definition of $D_g$ (the diameter of the gradient space), remember the local training step is operated on each client with different sparsity and different sparse masks, the corresponding computed gradient is thus bounded by $\beta_k D_g$. Again by the definition of $\beta$, which is the proportion of the remaining parameters of the aggregated global model $W$, we can thus further bound the gradient computed on each client by $\beta D_g$. 

Then we have
\begin{equation}
	\Vert \mathbf{v} \Vert<\frac{1}{\tau} \beta D_g.
\end{equation}

On the other hand, since $\langle \mathbf{v}, W'\rangle\sim \mathcal{N}(0,\Vert \mathbf{v}\Vert^2\sigma^2)$,
by Chernoff Bound technique, we have
\begin{equation}
   	\begin{aligned}
	\mathbb P\left(\langle \mathbf{v}, W'\rangle\ge \frac{\sqrt{2}\beta D_g\sigma}{\tau}\sqrt{\log\frac{1}{\delta}}\right)& \le	\mathbb P\left(\langle \mathbf{v}, W'\rangle\ge \sqrt{2}\Vert \mathbf{v}\Vert\sigma\sqrt{\log\frac{1}{\delta}}\right)\\
	&\le \min_{t}e^{-\sqrt{2}t\Vert \mathbf{v}\Vert\sigma\sqrt{\log\frac{1}{\delta}}}\mathbb{E}(e^{t\langle \mathbf{v}, W'\rangle}).
	\end{aligned} 
\end{equation}

For brevity, we define 
\begin{equation} \label{delta}
    \delta = \min_{t}e^{-\sqrt{2}t\Vert \mathbf{v}\Vert\sigma\sqrt{\log\frac{1}{\delta}}}\mathbb{E}(e^{t\langle \mathbf{v}, W'\rangle}).
\end{equation}
Therefore, with probability at least $1-\delta$ with respect to $W'$, we have that 
	\begin{equation}
	D_p(W') \le \frac{\sqrt{2}\beta D_g\sigma\frac{1}{\tau}\sqrt{\log\frac{1}{\delta}}+\frac{1}{\tau^2}\beta D_g^2}{2\sigma^2}.
	\end{equation}

Combining Lemma \ref{lemma:one}, we can have that the each step in Algorithm \ref{alg:algo1} is ($\tilde{\varepsilon},\frac{\tau}{N}\delta$)-differentially private, where $\tilde\varepsilon$ is defined as
\begin{equation}
	\tilde\varepsilon =\log \left(\frac{N-\tau}{N}+\frac{\tau}{N}\exp\left(\frac{\sqrt{2}\beta D_g\sigma\frac{1}{\tau}\sqrt{\log\frac{1}{\delta}}+\frac{1}{\tau^2}\beta^2D_g^2}{2\sigma^2}\right)\right).
\end{equation}

Applying Lemma \ref{lemma:multi}, we can conclude the differentially private guarantee ($\varepsilon',\delta'$) for the iterative steps.

Finally, combining Lemma \ref{lemma:g-dp} with ($\varepsilon',\delta'$) finish the proof.
\end{proof}

\end{document}